\documentclass[11pt,a4paper]{article}
\usepackage{times,latexsym}
\usepackage{url}
\usepackage[T1]{fontenc}

\usepackage[acceptedWithA]{tacl2018v2}   

\definecolor{applegreen}{rgb}{0.55, 0.71, 0.0}
\definecolor{Gray}{gray}{0.9}

\usepackage{soul}
\DeclareRobustCommand{\hlgray}[1]{{\sethlcolor{Gray}\hl{#1}}}

\usepackage{amsmath}
\usepackage{amssymb}
\usepackage{multirow}
\usepackage{url}
\usepackage{stmaryrd}
\usepackage{qtree}
\usepackage{tikz-qtree}
\usepackage{graphicx}
\usetikzlibrary{matrix}
\usetikzlibrary{fadings}
\usetikzlibrary{chains}
\usetikzlibrary{arrows}
\usetikzlibrary{shapes}
\usetikzlibrary{automata}
\usetikzlibrary{backgrounds}
\usetikzlibrary{decorations.pathreplacing,shapes,backgrounds}
\usetikzlibrary{fit}
\usetikzlibrary{patterns}
\usepackage{array}
\usepackage{arydshln}  
\usepackage{alltt}
\usepackage{listings}
\DeclareMathOperator*{\argmax}{arg\,max}

\usetikzlibrary{positioning,automata}

\usepackage{forest}

\newcommand{\added}[1]{\textcolor{black}{#1}}

\newcommand\T{\rule{0pt}{2.6ex}}       

\title{What Does My QA Model Know? \\ Devising Controlled Probes using Expert Knowledge}

\author{Kyle Richardson \and Ashish Sabharwal \\
  \ \\
  Allen Institute for AI, Seattle, WA, USA \\
  \texttt{\small \{kyler,ashishs\}@allenai.org} 
}
\date{}

\begin{document}

\maketitle

\begin{abstract}
Open-domain question answering (QA) involves many knowledge and reasoning challenges, but are successful QA models actually learning such knowledge when trained on benchmark QA tasks? We investigate this via several \emph{new diagnostic tasks} probing whether multiple-choice QA models know definitions and taxonomic reasoning---two skills widespread in existing benchmarks and fundamental to more complex reasoning. We introduce a methodology for automatically building probe datasets from \emph{expert knowledge sources}, allowing for systematic control and a comprehensive evaluation. We include ways to carefully control for artifacts that may arise during this process. Our evaluation confirms that transformer-based multiple-choice QA models are already predisposed to recognize certain types of structural linguistic knowledge. However, it also reveals a more nuanced picture: their performance notably degrades even with a slight increase in the number of ``hops'' in the underlying taxonomic hierarchy, and with more challenging distractor candidates. Further, existing models are far from perfect when assessed at the level of clusters of semantically connected probes, such as all hypernym questions about a single concept.
\end{abstract}

\section{Introduction}
\label{sec:intro}

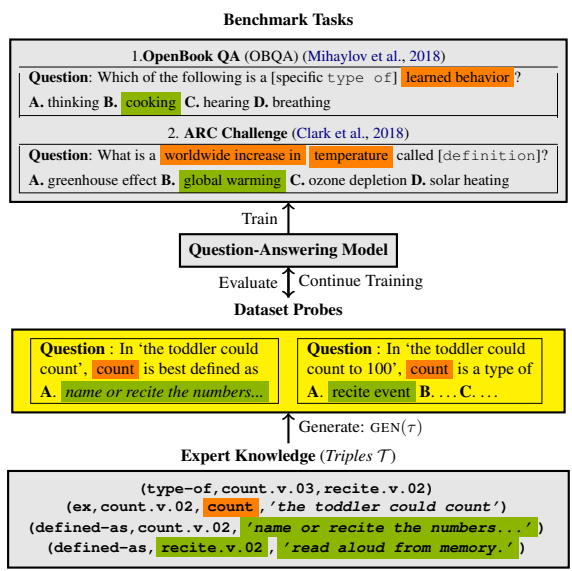
\begin{figure}[t]
    \centering
        \begin{tikzpicture}[scale=0.65, every node/.style={scale=0.58}]
        \tikzstyle{source} = [draw,thick,inner sep=.2cm,fill=Gray];
        \tikzstyle{source2} = [draw,thick,inner sep=.2cm,fill=yellow];
        \matrix (compiler) 
            [matrix of nodes,
                row sep=1cm,
                column sep=.8cm,
                ampersand replacement=\&,
                nodes={align=left}
                ]{
            \node[] (heading) {\textbf{Benchmark Tasks}}; \\[-.9cm]
            \node[source](examples) {\small
              \begin{tabular}{| l l |}
                 \multicolumn{2}{c}{1.\textbf{OpenBook QA} (OBQA) \cite{mihaylov2018can}} \\
                 \hline
                 \multicolumn{2}{|l|}{\textbf{Question}: Which of the following is a [specific \texttt{type of}] \colorbox{orange}{learned behavior}?} \\[.1cm]
                \multicolumn{2}{|l|}{\textbf{A.} thinking \textbf{B.} \colorbox{applegreen}{cooking} \textbf{C.} hearing \textbf{D.} breathing} \\ \hline
                 \multicolumn{2}{c}{} \\[-.2cm]
                 \multicolumn{2}{c}{2. \textbf{ARC Challenge} \cite{clark2018think}} \\
                 \hline
                 \multicolumn{2}{|l|}{\textbf{Question}: What is a \colorbox{orange}{worldwide increase in} \colorbox{orange}{temperature} called [\texttt{definition}]?} \\[.1cm]
                \multicolumn{2}{|l|}{\textbf{A.} greenhouse effect \textbf{B.} \colorbox{applegreen}{global warming} \textbf{C.} ozone depletion \textbf{D.} solar heating} \\
                 \hline

             \end{tabular}}; \\[-.6cm]
             \node[source] (model) {\textbf{Question-Answering Model}}; \\[-.6cm]
             \node[] (probes) {\textbf{Dataset Probes}}; \\[-.9cm]
             \node[source2] (synthetic) {
                \begin{tabular}{ c c c }
                {
                \begin{tabular}{| l | }
                \hline
                \textbf{Question} : In `the toddler could  \\ 
                count', \colorbox{orange}{count} is best defined as \\ 
                \textbf{A}. \colorbox{applegreen}{\emph{name or recite the numbers...}} \\ \hline
                \end{tabular}} & 
                {
                \begin{tabular}{| l | }
                \hline
                \textbf{Question} : In `the toddler could  \\ 
                count to 100', \colorbox{orange}{count} is a type of \\ 
                \textbf{A}. \colorbox{applegreen}{recite event}
                \textbf{B}. \dots
                \textbf{C}. \dots
                \\ \hline
                \end{tabular}}
                \\ 
                \end{tabular}
             }; \\[-.6cm]
             \node[] (knowledge) {\textbf{Expert Knowledge} (\emph{Triples} $\mathcal{T}$)}; \\[-1cm]
             \node[source](taxonomy) {\small
             \begin{tabular}{c}
              \textbf{\texttt{(type-of,count.v.03,recite.v.02)}} \\
              \textbf{\texttt{(ex,count.v.02,\colorbox{orange}{count},\emph{'the toddler could count'})}} \\ 
              \textbf{\texttt{(defined-as,\texttt{count.v.02},\emph{\colorbox{applegreen}{'name or recite the numbers...'}})}} \\ 
              \textbf{\texttt{(defined-as,\colorbox{applegreen}{recite.v.02},\colorbox{applegreen}{\emph{'read aloud from memory.'}})}} 
             \end{tabular}
             };  \\            
            };
            \draw[->,thick] (model) -- node[left] {Train\ \ } (examples);
            \draw[->,thick] (model) -- node[left] {Evaluate\ \ } (probes);
            \draw[->,thick] (probes) -- node[right] {\ Continue Training} (model);
            \draw[->,thick] (knowledge) -- node[right] {\ Generate: $\textsc{gen}(\tau)$} (synthetic); 
        \end{tikzpicture}
    \vspace{-4.5ex}
    \caption{An illustration of our experimental setup and probing methodology. The gray box at the top shows questions from existing open-domain QA benchmarks, requiring background knowledge. The yellow box shows simple examples of multiple-choice questions in our proposed Definition and \textit{ISA} probes.
    \vspace{-2.5ex}}
    \label{fig:mcqa}
\end{figure}

Automatically answering questions, especially in the open-domain setting where minimal or no contextual knowledge is explicitly provided, requires considerable background knowledge and reasoning abilities. For example, answering the two questions in the top gray box in Figure~\ref{fig:mcqa} requires identifying a specific \emph{ISA relation} (that `cooking' is a type of `learned behavior') as well as recalling a concept \emph{definition} (that `global warming' is defined as a `worldwide increase in temperature').


Recent success in QA has been driven largely by new benchmarks~\citep[etc.]{zellers2018swag,talmor2018commonsenseqa,bhagavatula2019abductive,khot2020qasc} and advances in model pre-training \cite{radford2018improving,devlin2018bert}. This raises a natural question: \emph{Do state-of-the-art multiple-choice QA (MCQA) models that excel at standard benchmarks truly possess basic knowledge and reasoning skills expected in these tasks?}

Answering this question is challenging due to limited understanding of heavily pre-trained complex models and the way existing MCQA datasets are constructed. We focus on the second aspect, which has two limitations: Large-scale crowdsourcing leaves little systematic control over question semantics or requisite background knowledge~\cite{welbl2017crowdsourcing}, while questions from real exams tend to mix multiple challenges in a single dataset, often even in a single question~\cite{clark2018think,boratko2018systematic}. 

To address this challenge, we propose systematically constructing model competence probes by exploiting structured information contained in \emph{expert knowledge sources} such as knowledge graphs and lexical taxonomies. Importantly, these probes are diagnostic tasks, designed not to impart new knowledge but to assess what models trained on standard QA benchmarks already know; as such, they serve as proxies for the types of questions that a model might encounter in its original task, but involve a single category of knowledge under various controlled conditions and perturbations.

Figure~\ref{fig:mcqa} illustrates our methodology. We start with a set of standard MCQA benchmark tasks $\mathcal{D}$ and a set of models $\mathcal{M}$ trained on $\mathcal{D}$. Our goal is to assess how competent these models are relative to a particular knowledge or reasoning skill $S$ (e.g., definitions) that is generally deemed important for performing well on $D$. To this end, we systematically and automatically generate a set of \emph{dataset probes} $P_S$ from information available in expert knowledge sources. Each probe is an MCQA rendering of the target information (see examples in Figure~\ref{fig:mcqa}, yellow box). We then use these probes $P_S$ to ask two empirical questions: (1) How well do models in $\mathcal{M}$ already trained on $\mathcal{D}$ perform on probing tasks $P_S$? (2) With additional nudging, can models be re-trained, using only a modest amount of additional data, to perform well on each probing task $P_{S}$ with minimal performance loss on their original tasks $D$ (thus giving evidence of prior model competence on $S$)? 

While our methodology is general, our experiments focus on probing state-of-the-art MCQA models in the domain of grade-school level science, which is considered particularly challenging with respect to background knowledge and inference~\cite{clark2015elementary,clark2019f,khot2020qasc}.
In addition, existing science benchmarks are known to involve widespread use of definition and taxonomic knowledge (see detailed analysis by \citet{clark2018think,boratko2018systematic}), which is also fundamental to deeper reasoning.
Accordingly, we employ the most widely used lexical ontology WordNet~\cite{miller1995wordnet} and publicly available dictionaries as sources of expert knowledge to construct our probes, WordNetQA (Section~\ref{sec:wordnetqa}) and DictionaryQA (Section~\ref{sec:dqa})\footnote{\added{All data and code are available at \url{https://github.com/allenai/semantic_fragments}}}. These probes measure competence in various settings including hypernymy, hyponymy, and synonymy detection, as well as word sense disambiguation.   

Our exploration is closely related to the recent work of \citet{talmor2019olmpics}. However, a key difference is that they study language models (LMs), for which there is \emph{no clear \textit{a priori} expectation} of specific knowledge or reasoning skills. In contrast, we focus on models heavily trained for benchmark QA tasks, where such tasks are known to require certain types of knowledge and reasoning skills. We probe whether such skills are actually learned by QA models, either during LM pre-training or when training for the QA tasks.

Recognizing the need for suitable controls in any synthetic probing methodology~\cite{hewitt2019designing,talmor2019olmpics}, we introduce two controls: (a)
the probe must be challenging for any model that lacks contextual embeddings, and (b) strong models must have a \emph{low inoculation cost}, i.e., when fine-tuned on a few probing examples, the model should mostly retain its performance on its original task.\footnote{Standard inoculation~\citep{liu2019inoculation} is known to drop performance on the original task. We use a modified objective~\citep{richardson2020probing} to alleviate this issue.} This ensures that the probe performance of a model, even when lightly inoculated on probing data, reflects its knowledge as originally trained for the benchmark task, which is precisely what we aim to uncover.

Constructing a wide range of systematic tests is critical for having definitive empirical evidence of model competence on any given phenomenon. Such tests should cover a broad set of concepts and question \emph{variations} (i.e., systematic adjustments to how the questions are constructed). When assessing \emph{ISA} reasoning, not only is it important to recognize in the question in Figure~\ref{fig:mcqa} that \emph{cooking} is a \emph{learned behavior}, but also that \emph{cooking} is a general type of \emph{behavior} or, through a few more inferential steps, a type of \emph{human activity}. Our automatic use of expert knowledge sources allows constructing such high-coverage probes, circumventing pitfalls of solicitation bias and reporting bias.

Our results confirm that transformer-based QA models\footnote{Different from \citet{talmor2019olmpics}, we find BERT and RoBERTa based QA models to be qualitatively similar, performing within 5\% of each other on nearly all probes.} have a remarkable ability to recognize the types of knowledge captured in our probes---even without additional fine-tuning (i.e., in a \emph{zero-shot} setting). Such models can even outperform strong task-specific non-transformer models trained directly on our probing tasks
(e.g., +26\% compared to a task-specific LSTM).
We also show that the same models can be effectively re-fine-tuned on small samples (even 100 examples) of probe data, and that high performance on the probes tends to correlate with a smaller drop in the model's performance on the original QA task. 

Our comprehensive assessment also reveals important nuances to the positive trend. For example, we find that the best models still perform 2-10\% (absolute) below conservative estimates of human performance (Section~\ref{sec:human-performance}) on these tasks. Further, the accuracy of even the best QA model degrades substantially on our hyponym probes (by 8-15\%) when going from 1-hop hyponym links to 2-hops. The accuracy on the WordNetQA probe drops by 14-44\% under our \emph{cluster-level analysis} (Section~\ref{sec:clusters}), which assesses whether a model knows several facts about each individual concept, rather than only answering correctly isolated questions. This shows that state-of-the-art QA models have much room to improve even in some fundamental building blocks (definitions and taxonomic hierarchies) of more complex forms of reasoning.

\section{Related Work}

We follow recent work on constructing challenge datasets for probing neural models, which has primarily focused on the task of natural language inference (NLI) \cite{glockner2018breaking,mccoy2019right,rozen2019diversify,warstadt2019investigating}. Most of this work looks at constructing data through adversarial generation methods, \added{which have also been found useful for creating stronger models \cite{Kang2018AdvEntuReAT}.} There has also been work on using synthetic data of the type we consider in this paper \cite{poliak2018collecting,geiger2019posing,yanaka2020neural,clark2020transformers}. We closely follow the methodology of \citet{richardson2020probing}, who use hand-constructed linguistic fragments to probe NLI models and study model re-training using a variant of the \emph{inoculation by fine-tuning} strategy of \citet{liu2019inoculation}. In contrast, we focus on probing open-domain MCQA models (see \citet{si2019does} for a study on \emph{reading comprehension}) as well as constructing data from much larger sources of structured knowledge.  

Our main study focuses on probing the BERT model and fine-tuning approach of \citet{devlin2018bert}, and other variants thereof, which are all based on the transformer architecture of \citet{vaswani2017attention}. There have been recent studies into the types of relational knowledge contained in large-scale knowledge models \cite{schick2019rare,petroni2019language,jiang2019can}, which also probe models using structured knowledge sources. These studies, however, primarily focus on unearthing the knowledge contained in the underlying language models \emph{as is} without further training, using simple (single token) cloze-style probing tasks and templates. Most of these results only provide a \emph{lower-bound} estimate of model performance, since the probing templates being employed potentially deviate from what the model has observed during pre-training. In contrast, we focus on understanding the knowledge contained in language models \emph{after} they have been trained for a QA end-task using benchmark datasets in which such knowledge is expected to be widespread. Further, our evaluation is done before and \emph{after} these models are fine-tuned on our small samples of target data. This has the advantage of allowing each model to become informed about the format of each probe. We also explore a more complex set of probing templates.


The use of lexical resources such as WordNet to construct datasets has a long history, and has recently appeared in work on adversarial attacks \cite{jia2017adversarial} and general task construction \cite{pilehvar2019wic}. In the area of MCQA, there is related work on constructing questions from tuples \cite{jauhar2016tables,talmor2018commonsenseqa}, both of which involve standard crowd annotation to elicit question-answer pairs (see also \citet{seyler2017knowledge,reddy2017generating}). In contrast to this work, we focus on generating data in an entirely automatic and \emph{silver-standard} fashion (i.e., in a way that potentially introduces a little noise), which obviates the need for expensive annotation and gives us the flexibility to construct much larger datasets that control a rich set of semantic aspects of the target questions. Following standard practices in MCQA dataset creation~\citep[e.g.,][]{khot2020qasc}, however, we perform crowd-sourcing to obtain \emph{conservative} (in the sense of \citet{nangia2019human}) estimates of human performance on our main evaluation sets, to compare against model performance.

While our probing methodology is amenable to any domain, we focus on probing open-domain QA models in the domain on grade-school level science using a standard suite of benchmark QA datasets (see Table~\ref{tab:benchmarks}). Our choice of this domain is based on the following considerations: it is well-studied qualitatively \cite{davis2016write}, making it relatively easy to know the types of probes and diagnostic tests to construct using existing expert knowledge. For example, the manual analysis of \citet{mihaylov2018can}  found that \emph{explicit} definitional and ISA knowledge occurred in around 20\% and 18\%, respectively, of the questions sampled in one benchmark task. \citet{clark2013study} and \citet{boratko2018systematic} provide similar results involving other benchmarks used in our study.

We also examined MCQA models trained on closely-related datasets tailored to commonsense and situational reasoning \cite{zellers2018swag,talmor2018commonsenseqa,bhagavatula2019abductive,sap2019socialiqa}. However, there has been a limited study of the kinds of knowledge needed in this domain, as well as expert knowledge sources for creating corresponding probes. MCQA models trained in this domain exhibit lower performance on our definition and ISA probes.

\section{Dataset Probes and Construction} 
\label{sec:probes}

\begin{table}[t]
    \centering
    \scalebox{0.7}{
        \begin{tabular}{| c | l  l |}
            \hline 
            \multicolumn{1}{|c}{\textbf{Set}} & \multicolumn{1}{c}{\textbf{WordNet} (WN)} & \multicolumn{1}{c|}{\textbf{GCIDE}}    \\ \hline 
            $\mathcal{R}$ & \texttt{\{isa$\uparrow$,isa$\downarrow$,} & \texttt{\{def, ex, lemma\}} \\ 
            & \texttt{ def, ex, lemma\}} & \\
            $\mathcal{C}$ & \texttt{\{\textbf{WN synsets}\}} & \texttt{\{\textbf{entry ids}\}} \\ 
            $\mathcal{D}$ & \texttt{\{\textbf{synset glosses}\}} & \texttt{\{\textbf{unique defs}\}} \\ 
            $\mathcal{S}$ & \texttt{\{\textbf{synset sentences}\}} & \texttt{\{\textbf{entry examples}\}} \\ 
            $\mathcal{W}$ & \texttt{\{\textbf{synset lemmas}\}} & \texttt{\{\textbf{all words}\}} \\ \hline \hline  
            \multicolumn{2}{|c}{\textbf{Atomic Triple Types}} & \multicolumn{1}{c|}{\textbf{Definition}} \\ \hline 
            \multicolumn{2}{|l}{Concept Senses and Definitions} & \multicolumn{1}{l|}{$\mathcal{T}_{d} \subseteq \texttt{\{def\}} \times \mathcal{C} \times \mathcal{D}$} \\
            \multicolumn{2}{|l}{Concepts with Example Sentences} & \multicolumn{1}{l|}{$\mathcal{T}_{e} \subseteq \texttt{\{ext\}} \times \mathcal{C} \times \mathcal{S}$} \\ 
            \multicolumn{2}{|l}{Concepts with Words} & \multicolumn{1}{l|}{$\mathcal{T}_{l} \subseteq \texttt{\{lemma\}} \times \mathcal{C} \times \mathcal{W}$} \\ 
            \multicolumn{2}{|l}{ISA Relations (\emph{WN only})} & \multicolumn{1}{l|}{$\mathcal{T}_{i} \subseteq \texttt{\{isa$^\uparrow$,isa$^\downarrow$\}} \times \mathcal{C} \times \mathcal{C}$} \\ \hline 
        \end{tabular}}
    \caption{A description of the different resources used to construct the probes, represented as abstract triples.}
    \label{fig:resources}
\end{table}

Our probing methodology starts by constructing challenge datasets (Figure~\ref{fig:mcqa}, yellow box) from a target set of knowledge resources. Each probing dataset consists of multiple-choice questions that include a \emph{question} $\textbf{q}$ and a set of \emph{answer choices} or candidates $\{a_{1},...a_{N}\}$. This section describes in detail the 5 datasets we build (grouped into \textbf{WordNetQA} and \textbf{DictionaryQA}), drawn from two publicly-available resources: WordNet \cite{miller1995wordnet} and the GNU Collaborative International Dictionary of English (GCIDE).\footnote{see \url{https://wordnet.princeton.edu/} and \url{http://gcide.gnu.org.ua/}}

For convenience, we will describe each source of expert knowledge as a directed, edge-labeled graph $G$. The nodes of this graph are $\mathcal{V} = \mathcal{C} \cup \mathcal{W} \cup \mathcal{S} \cup \mathcal{D}$, where $\mathcal{C}$ is a set of atomic concepts, $\mathcal{W}$ a set of words, $\mathcal{S}$ a set of sentences, and $\mathcal{D}$ a set of definitions (see Table~\ref{fig:resources} for details for WordNet and GCIDE). Each edge of $G$ is directed from an atomic concept in $\mathcal{C}$ to another node in $V$, and is labeled with a relation, such as hypernym or \texttt{isa$^\uparrow$}, from a set of relations $\mathcal{R}$ (see Table~\ref{fig:resources}).

When defining our probe question templates, it will be useful to view $G$ as a set of \textit{(relation, source, target)} \textbf{triples} $\mathcal{T} \subseteq \mathcal{R} \times \mathcal{C} \times \mathcal{V}$. Due to their origin in an expert knowledge source, such triples preserve semantic consistency. For instance, when the \textit{relation} in a triple is \texttt{def}, the corresponding edge maps a concept in $\mathcal{C}$ to a definition in $\mathcal{D}$.


\begin{figure}
\newdimen\nodeDist
\nodeDist=32mm
    \centering
        \begin{tikzpicture}[
    node/.style={%
      draw,
    },
    scale=0.6, 
    every node/.style={scale=0.6}]
  ]
    \node [node,fill=applegreen] (A) {\texttt{utter.v.01}};
    \path (A) ++(-135:\nodeDist) node [node,fill=Gray] (B) {\texttt{parrot.v.02}};
    \path (A) ++(-45:\nodeDist) node [node,rectangle,fill=applegreen] (C) {\texttt{recite.v.02}};
    \path (C) ++(-135:\nodeDist) node [node,fill=Gray] (D) {\texttt{spell.v.01}};
    \path (C) ++(-45:\nodeDist) node [node,fill=white] (E) {\texttt{count.v.03}};
    \path (E) ++(-90:17mm) node [node,fill=Gray] (F) {\texttt{count-down.v.01}};
    \path (D) ++(-90:17mm) node [node,fill=Gray] (G) {\texttt{mispell.v.01}};

    \draw[<-,line width=0.50mm,draw=red] (A) -- (B) node [left,pos=0.25] {\texttt{ISA}$^{\uparrow}$}(A);
    \draw[<-,line width=0.50mm,draw=applegreen] (A) -- (C) node [right,pos=0.25] {\texttt{ISA}$^{\uparrow}$}(A);
    \draw[<-,line width=0.50mm,draw=red] (C) -- (D) node [left,pos=0.25] {\texttt{ISA}$^{\uparrow}$}(A);
    \draw[<-,line width=0.50mm,draw=applegreen] (C) -- (E) node [right,pos=0.25] {\texttt{ISA}$^{\uparrow}$}(A);
    \draw[<-,line width=0.50mm,draw=red] (D) -- (G) node [right,pos=0.25] {\texttt{ISA}$^{\uparrow}$}(A);
    \draw[<-,line width=0.50mm,draw=red] (E) -- (F) node [right,pos=0.25] {\texttt{ISA}$^{\uparrow}$}(A);
\end{tikzpicture}
    \\[.4cm]
    {\scriptsize
        \begin{tabular}{| p{22mm} p{42mm} |}
            \hline 
            \multicolumn{1}{|c}{\textbf{Graph Triples}} & \multicolumn{1}{c|}{\textbf{Question/Answers}}  \\ \hline \hline 
            \multicolumn{2}{|c|}{\textbf{Question+Answer about Hypernymy/ISA$^{\uparrow}$}} \\ \hline 
             \texttt{(isa$^\uparrow$,} \texttt{count.v.03} \texttt{,recite.v.02)} & 
             $\textbf{q}.$ In the sentence \hlgray{The toddler could count}, the word \hlgray{count} is a type of: $\textbf{a}.$ \hlgray{recite event}... \\ \hline 
             \multicolumn{2}{|c|}{\textbf{Sister Family Distractors}} \\ \hline
             \texttt{(isa$^\downarrow$,} \texttt{recite.v.02} \texttt{,spell.v.01)} & $a'_{1}$. \hlgray{spell event}, defined as ... (\emph{1-hop sister distractor}); 
             $a'_{2}$ \hlgray{mispell event}, defined as... (\emph{2-hop sister}).
             \\ \hline 
        \end{tabular}}
    \caption{A portion of the WordNet \texttt{ISA} graph (top) and an example distractor function \textsc{Distr}$(\tau)$ (bottom)  used to generate distractor choices $\{a'_{1},a'_{2}\}$ for a question $\textbf{q}$ based on information in the graph.}
    \label{fig:distractor_generator}
\end{figure}


\begin{table*}[t]
    \centering
        {\scriptsize
        \begin{tabular}{| p{20mm} | p{20mm} | p{37mm} | p{55mm} |}
        
            \hline 
             \textbf{Probe Type} & \multicolumn{1}{c|}{\textbf{Triple Input }$\tau$} &  \multicolumn{1}{c|}{\textbf{Generation Templates} from $\mathcal{Q}$} & \multicolumn{1}{c|}{\textbf{Example Questions and Answers} $(q,a)$} \\ \hline 
             \texttt{Definitions}: Defining words in context. & 
                                    $(\texttt{def},c_{i},d)$ $(\texttt{ex},c_{i},s)$ $(\texttt{word},c_{i},w)$
                                    & $\textbf{q}.$\emph{ In the sentence \hlgray{$[s]$}, the word \hlgray{$[w]$} is best defined as: $\textbf{a}.$ \hlgray{$[d]$}} & 
                                     $\textbf{q}.$ In the sentence \hlgray{The baby nestled her head}, the word \hlgray{nestled} is best defined as:  $\textbf{a}.$ \hlgray{position comfortably}
                                    \\ \hline 
             \texttt{Hypernymy}: ISA$^{\uparrow}$ reasoning in context (symbolically \texttt{$c_{i}$=>$c_{i'}$}). & 
                                            $(\texttt{def},c_{i'},d)$ $(\texttt{isa}^\uparrow,c_{i},c_{i'})$ $(\texttt{ex},c_{i},s)$ $(\texttt{word},c_{i},w)$ $(\texttt{word},c_{i'},w')$
                                            & $\textbf{q}.$\emph{ In \hlgray{$[s]$}, the word or concept \hlgray{$[w]$} is best described as a type of $\textbf{a}.$  \hlgray{$[w']$} defined as  \hlgray{$[d]$}} 
                                            & $\textbf{q}.$ In \hlgray{The thief eluded the police}, the word or concept \hlgray{eluded} is best described as a type of  $\textbf{a}.$ \hlgray{escape event} defined as \hlgray{to run away from..}
                                            \\ \hline 
             \texttt{Hyponymy}: ISA$^{\downarrow}$ reasoning given context. (symbolically \texttt{$c_{i}$<=$c_{i'}$}) & 
                                                    $(\texttt{def},c_{i'},d)$ $(\texttt{isa}^{\downarrow},c_{i},c_{i'})$ $(\texttt{ex},c_{i},s)$ $(\texttt{word},c_{i},w)$ $(\texttt{word},c_{i},w')$
                                                    & $\textbf{q}.$\emph{ Given the context \hlgray{$[s]$}, which of the following word or concept is a specific type of  \hlgray{$[w]$} $\textbf{a}.$ \hlgray{$[w']$} defined as  \hlgray{$[d]$}} 
                                                    & $\textbf{q}.$ Given the context \hlgray{they awaited her arrival}, which of the following word or concept is a specific type of \hlgray{arrival}? $\textbf{a}.$ \hlgray{crash landing}, defined as \hlgray{an emergency landing under circumstances where...}.'
                                                    \\ \hline 
             \texttt{Synonymy}: Related words. & 
                               $(\texttt{def},c_{i},d)$  $(\texttt{word},c_{i},w_{1})$ $(\texttt{word},c_{i},w_{2})$
                               & $\textbf{q}.$\emph{ Which words best correspond to \hlgray{$[d]$}? $\textbf{a}.$ \hlgray{$[\{w_{1},w_{2},...\}]$}}
                               & $\textbf{q}.$ Which set of words best corresponds to the definition \hlgray{a grammatical category in inflected languages governing} \hlgray{agreement ....}? $\textbf{a}.$ \hlgray{gender,...}
                               \\ \hline 
        \end{tabular}}
    \caption{Details of the \textsc{gen}$(\tau)$ function used to construct gold question-answer pairs $(\textbf{q},\textbf{a})$ from a triple graph $G$.}
    \label{table:wordnet_gen}
\end{table*}

We rely on two heuristic functions, defined below for each individual probe: $\textsc{gen}_{\mathcal{Q}}(\tau)$, which generates gold question-answer pairs $(\textbf{q},\textbf{a})$ from a set of triples $\tau \subseteq \mathcal{T}$ and question templates $\mathcal{Q}$, and $\textsc{distr}(\tau')$, which generates distractor answers choices $\{ a'_{1},...a'_{N-1} \}$ based on another set of triples $\tau'$ (where usually $\tau \subset \tau'$). For brevity, we will use $\textsc{gen}(\tau)$ to denote $\textsc{gen}_{\mathcal{Q}}(\tau)$.

In generating our dataset probes, our general strategy is to build automatic \emph{silver-standard} training  and developments sets, in the latter case at a large scale to facilitate detailed and controlled analysis of model performance. As discussed below, we also provide estimates of human performance on our test sets, and in some cases introduce smaller gold-standard test sets to allow for a direct comparison with model performance.

\subsection{WordNetQA} 
\label{sec:wordnetqa}

WordNet is a publicly-available English lexical database consisting of around 117k concepts, which are organized into groups of \emph{synsets} that each contain a \emph{gloss} (i.e., a definition), a set of representative English words (called \emph{lemmas}), and, in around 33k synsets, example sentences. In addition, many synsets have \texttt{ISA} links to other synsets that express complex taxonomic relations. Figure~\ref{fig:distractor_generator} shows an example and Table~\ref{fig:resources} summarizes how we formulate WordNet as a set of triples $\mathcal{T}$ of various types. These triples together represent a directed, edge-labeled graph $G$.

Our main motivation for using WordNet, as opposed to a resource such as ConceptNet \cite{havasi2007conceptnet}, is the availability of glosses ($\mathcal{D}$) and example sentences ($\mathcal{S}$), which allows us to construct natural language questions that contextualize  the types of concepts we want to probe. For example, when probing whether a model has knowledge of a concept such as \emph{bank} (a financial institution), we provide an example sentence \emph{he cashed a check at the bank}, to help disambiguate the particular sense of \emph{bank} we are probing. Sentential contexts also provide additional hints to models in cases of rare or infrequent concepts.\footnote{Given the open-domain nature of WordNet, not all probed concepts may have \emph{explicitly} been observed during QA training. Nevertheless, unlike prior probing studies \cite{petroni2019language}, we did not see a substantial performance disparity between observed and unobserved concepts across our models, perhaps owing to the provided contexts.} Since WordNet is the most authoritative and widely-used knowledge resource in NLP, it also has the advantage of having mappings into other knowledge resources \cite{niles2001towards,navigli2010babelnet,tandon2017webchild}, which allows for easily extending our probes to other domains and phenomena.

\paragraph{Example Generation $\textsc{gen}(\tau)$.}
We build 4 individual datasets based on semantic relations native to WordNet: \texttt{hypernymy} (i.e., generalization or ISA reasoning up a taxonomy, ISA$^\uparrow$), \texttt{hyponymy} (ISA$^{\downarrow}$), \texttt{synonymy}, and \texttt{definitions}. To generate a set of questions in each case, we employ a number of rule templates $\mathcal{Q}$ that operate over tuples. A subset of such templates is shown in  Table~\ref{table:wordnet_gen} and were designed to mimic \emph{naturalistic} (i.e., human authored) questions we observed in our science benchmarks.

For example, suppose we wish to create a question $\textbf{q}$ about the definition of a target concept $c \in \mathcal{C}$. We first select a question template from $\mathcal{Q}$ that first introduces the concept $c$ and its lemma $l \in \mathcal{W}$ in context using the example sentence context $s \in \mathcal{S}$, and then asks to identify the corresponding WordNet gloss $d \in \mathcal{D}$, which serves as the gold answer $\textbf{a}$. The same is done for ISA reasoning; each question about a hypernym/hyponym relation between two concepts $c \to^{\uparrow/\downarrow} c' \in \mathcal{T}_{i}$ (e.g., $\texttt{dog} \to^{\uparrow/\downarrow} \texttt{animal/terrier}$) first introduces a context for $c$ and then asks for an answer that identifies $c'$ (which is also provided with a gloss so as to contain all available context).


\begin{table*}[t]
    \centering
    \scalebox{0.72}{
        \begin{tabular}{| p{4.3cm} | p{7.5cm} | p{8.6cm} |}
        \hline 
        \multicolumn{1}{|c|}{\textbf{Target Concept}} & \multicolumn{1}{c|}{\textbf{Example Question}} & \multicolumn{1}{c|}{\textbf{Inferences}}  \\ 
        & & \multicolumn{1}{c|}{\emph{(target answers in symbolic form)}} \\ \hline  
        \textbf{trouser.n.01}, \textbf{gloss}: \emph{a garment extending from the waist to the knee or ankle covering each leg..} & 
        $\textbf{q}.$ In \hlgray{he had a sharp crease in his trousers}, the word/phrase \hlgray{trousers} is best defined as a type of &
             \texttt{trouser.n.01 => \textbf{consumer\_goods.n.01}} 
             \texttt{trouser.n.01 => \textbf{garment.n.01}}
             \texttt{trouser.n.01 => \textbf{commodity.n.01}}
             \texttt{trouser.n.01 => \textbf{clothing.n.01}}
        \\ \hline 
        \textbf{oppose.v.06}, \textbf{gloss}: \emph{be resistant to} &
        $\textbf{q}.$ In the sentence or expression \hlgray{The board opposed his motion}, the following is a more specific type of \hlgray{opposed} [or opposition] & 
            \texttt{oppose.v.06~<=~\textbf{protest.v.02}}
            \texttt{oppose.v.06~<=~\textbf{veto.v.01}}
            \texttt{oppose.v.06~<=~\textbf{demonstrate.v.04}}
        \\ \hline 
        \textbf{poet\_laureate.n.01}, \textbf{gloss}: \emph{a poet who is ... holding an honorary position...} & 
        $\textbf{q}.$ Given the fragment \hlgray{he is the poet laureate of Arkansas}, \hlgray{poet laureate} ... is best described as a type of & 
        \texttt{poet\_laureate.n.01=>\textbf{poet.n.01}}
        \texttt{poet\_laureate.n.01=>\textbf{communicator.n.01}}
        \texttt{poet\_laureate.n.01=>\textbf{writer.n.01}}
        \\ \hline 
         \end{tabular}}
    \caption{Semantic clusters for \added{three} target concepts, involving ISA reasoning.}
    \label{tab:clusters}
\end{table*}

In the latter case, the rules $(\texttt{isa}^{r},c,c') \in \mathcal{T}_i$ in Table~\ref{table:wordnet_gen} cover only \emph{direct} ISA links from $c$ in direction $r \in \{ \uparrow,\downarrow \}$. In practice, for each $c$ and $r$, we construct tests that cover the set \textsc{HOPS}$(c,r)$ of \emph{all} direct as well as derived ISA relations of $c$: 
\begin{align*}
    \textsc{hops}(c,r) \texttt{:=} \Big\{ (\texttt{isa}^{r},c,c') \in \mathcal{T}_{i} \Big\} \cup \textsc{hops}(c',r) 
\end{align*}
This allows us to evaluate the extent to which models are able to handle complex forms of reasoning that require several inferential steps or \emph{hops}.\footnote{In practice, most WordNet synsets have no more than 5 hops. We use this as a default limit when building datasets.} 

\paragraph{Distractor Generation: $\textsc{distr}(\tau')$.}
Figure~\ref{fig:distractor_generator} shows an example of how distractors are generated, relying on similar principles as above. For each concept $c$, we choose 4 distractor answers that are close in the WordNet semantic space. For example, when constructing hypernymy tests for $c$ from the set \textsc{hops}$(c,\uparrow)$, we draw distractors from $\textsc{hops}(c,\downarrow)$, as well as from the \emph{$\ell$-deep sister family} of $c$, defined as follows. The $1$-deep sister family is simply $c$'s siblings or sisters, i.e., the other children $\tilde{c} \neq c$ of the parent node $c'$ of $c$. For $\ell > 1$, the $\ell$-deep sister family also includes all descendants of each $\tilde{c}$ up to $\ell-1$ levels deep, denoted $\textsc{hops}_{\ell-1}(\tilde{c},\downarrow)$. Formally:
\begin{align*}
    \textsc{sister}_{\ell}(c)\, &\texttt{:=}\, \Big\{ x \in \textsc{hops}_{\ell-1}(\tilde{c},\downarrow) \mid \\
         &(\texttt{isa}^{\uparrow},c,c') \in \mathcal{T}_{i}, \\ 
         &(\texttt{isa}^{\uparrow},\tilde{c},c') \in \mathcal{T}_{i}, \, 
         \tilde{c} \neq c \Big\}  
\end{align*}

For definitions and synonyms, we build distractors from all of these sets (with a similar depth limit for \textsc{sister} distractors), enabling a systematic investigation via a wide range of distractors.

\subsubsection{Perturbations and Semantic Clusters}
\label{sec:clusters}
For each concept $c$ (an atomic WordNet synset) and probe type (definitions, hypernymy, etc.), we have a wide variety of questions related to $c$ that manipulate 1) the complexity of reasoning that is involved (e.g., the number of inferential \emph{hops}) and 2) the types of distractors (or \emph{distractor perturbations}) that are employed. We call such sets \emph{semantic clusters}. 

Table~\ref{tab:clusters} shows \added{three} examples, capturing ISA reasoning about the following target concepts: \emph{trousers}, \emph{opposing}, and \emph{poet laureate}. Such clusters enable new types of evaluation of the comprehensiveness and consistency of a model's knowledge of target concepts.

\subsubsection{Summary of Probe Datasets}
\label{sec:dataset-details}

Details of the individual datasets, including average cluster sizes, are summarized in Table~\ref{tab:wordnetqa}.


\begin{table}[ht]
\centering 
\scalebox{0.69}{
\begin{tabular}{ l c c c }
    \hline 
     \multicolumn{1}{l}{\textbf{Probe}} & \textbf{\# Questions} & \textbf{Cluster Size} & \textbf{\# Synsets}  \\ 
      & \emph{(Unique / w Perturb.)} & \emph{(Avg.)} & \emph{(or concepts)}  \\ \hline 
     \textbf{Hypernymy} & 19,705 / 35,094 &  5 & 7,849 \\ 
     \textbf{Hyponymy} & 6,697 / 35,243 &  11 & 3,452 \\
     \textbf{Synonymy} & 28,254 / 91,069 & 6 & 15,632 \\ 
     \textbf{Definitions} & 31,380 / 148,662 & 10 & 15,159 \\  \cdashline{1-4}
     \textbf{WordSense} & $\sim$7,000 / -- \ \ \ \ \ \ \ \ \ \ \ \ & 1 & \!\!\!$\sim$7,000 \\  \hline
\end{tabular}}

\caption{
    Details of our dataset probes, including both the number of \emph{unique} $(\textbf{q},\textbf{a})$ pairs (for \textbf{WordNetQA}) and the number of all questions including distractor choice perturbations (\emph{w Perturb.}).
}
\label{tab:wordnetqa}
\end{table}

From these sets, we follow \citet{richardson2020probing} in allocating a maximum of 3k examples for \emph{inoculating} the models in the manner described in the next section (i.e., for continuing to train QA models and introduce them to the format of our probes), and reserve the rest for development and testing. In particular, we build large development sets, which are important for performing detailed analysis and cluster-based evaluation. 

\subsubsection{Human Performance}
\label{sec:human-performance}

We report human scores on the individual test sets in WordNetQA (see bottom of Table~\ref{tab:instance}). This is done in two ways.

First, for our test sets generated for definitions and synonyms that cover a large set of disconnected concepts in the WordNet graph and where it is infeasible to annotate  individual instances of concepts, we estimate human performance by having crowd-workers on Amazon Mechanical Turk answer a random sample of 500 test questions. Scores are computed by taking the majority vote for each question among 5 annotators. This follows exactly the evaluation protocol employed by \citet{nangia2019human} and is a \emph{conservative} estimate in that crowd annotators received virtually no training and no qualification exam before participating in the task.

Second, for our hypernymy and hyponymy test sets, which cover a smaller number of densely-connected concepts, we annotated smaller \emph{gold-standard} test sets that include a sample of around 2,000 random questions that cover a large proportion of the concepts being probed and that have high human performance. To do this, we follow the annotation strategy described above, and greedily apply filtering to remove questions incorrectly answered by human annotators, which follows prior work on building evaluation sets for MCQA \cite{mihaylov2018can,talmor2018commonsenseqa,khot2020qasc}.

\subsection{DictionaryQA} 
\label{sec:dqa}


 
The DictionaryQA dataset is created from the English dictionary GCIDE built largely from the Webster's Revised Unabridged Dictionary \cite{webster1913webster}, which has previously been used in other NLP studies due to its large size and public availability~\cite{hill2016learning}.  Each dictionary entry consists of a word, its part-of-speech, its definition, and an optional example sentence, as shown for an example in Table~\ref{fig:gcide}.


\begin{table}[ht]
    \centering
    
    {\scriptsize 
        \begin{tabular}{| p{0.42\textwidth}  |}
            \hline 
            \multicolumn{1}{|c|}{\textbf{GCIDE Dictionary Entries}} \\ \hline\hline 
             \textbf{word}: gift, \textbf{pos}: n., \textbf{definition}: Anything given; anything voluntarily transferred by one person to another without compensation; a present; \textbf{entry example}: \textcolor{red}{None.} \\ \hline 
             \textbf{word}: gift, \textbf{pos}: n., \textbf{definition}: A bribe; anything given to corrupt. \textbf{entry example}: \textcolor{red}{None.} \\
             \hline
             \textbf{word}: gift, \textbf{pos}: n., \textbf{definition}: Some exception inborn quality or characteristic; a striking or special talent or aptitude;.. \textbf{entry example}: \emph{the gift of wit; a gift for speaking.} 
             \\ \hline 
        \end{tabular}}
    
    \caption{
        Example dictionary entries for the word \emph{gift}.
    }
    \label{fig:gcide}
\end{table}

Overall, 33k entries (out of a total of 155k) contain example sentences/usages. As with the WordNet probes, we focus on this subset so as to contextualize each word being probed. Since GCIDE does not have ISA relations or explicit synsets, we take each unique entry to be a distinct sense. Our probe centers around word-sense disambiguation.

%
To buildQA examples, we use the same generation templates for \emph{definitions} exemplified in Table~\ref{table:wordnet_gen} for WordNetQA. To construct distractors, we simply take alternative definitions for the target words that represent a different word sense (e.g., the alternative definitions of \emph{gift} in Table~\ref{fig:gcide}), and randomly chosen definitions if needed to create a 5-way multiple choice question. As above, we reserve a maximum of 3k examples for training, and use the same amount for development.

Our initial attempts at building this dataset via standard random splitting resulted in certain systematic biases, revealed by high performance of the \textbf{choice-only} model we used as a control. Among other factors, we found the use of definitions from entries without example sentences as distractors (see again Table~\ref{fig:gcide}) to have a surprising correlation with such biases. Filtering such distractors helped improve the quality of this probe.

For assessing human performance, we annotated a smaller gold-standard test set consisting of around 1,100 questions using the crowd-sourcing elicitation setup described in Section~\ref{sec:wordnetqa}.

\section{Probing Methodology and Modeling}

Given the probes above, we now can start to answer the empirical questions posed at the beginning. Our main focus is on looking at transformer-based MCQA models trained on science benchmarks in Table~\ref{tab:benchmarks}. We start with our target MCQA models, as well as several control baselines.

\subsection{Task Definition and Modeling}

Given a dataset $D =\{ (\textbf{q}^{(d)}, \{ a_{1}^{(d)},..., a_{N}^{(d)}\}) \}_{d}^{\mid D \mid}$ consisting of pairs of questions stems $\textbf{q}$ and answer choices $a_{i}$, the goal is to find the correct answer $a_{i^{*}}$ that correctly answers each $\textbf{q}$. Throughout this paper, we look at 5-way multiple-choice problems (i.e., where each $N=5$). 

\paragraph{Question+Answer Encoder.}
Our investigation centers around the use of the transformer-based BERT encoder and fine-tuning approach of \citet{devlin2018bert} (see also \citet{radford2018improving}). For each question and individual answer pair $q^{(j)}_{a_{i}}$, we assume the following rendering of this input:
\begin{align*}
{\footnotesize
    q^{(j)}_{a_{i}} := \texttt{[CLS] $\textbf{q}^{(j)}$ [SEP] $a^{(j)}_{i}$ [SEP]}}
\end{align*}
This is run through the pre-trained BERT encoder to generate a representation for $ q^{(j)}_{a_{i}}$ using the hidden state representation for \texttt{CLS} (i.e., the \emph{classifier token}):
$\textbf{c}^{(j)}_{i} = \textsc{BERT}(q^{(j)}_{a_{i}}) \in \mathbb{R}^{H}$.
The probability of a given answer $p^{(j)}_{i}$ is then standardly computed using an additional classification layer over $\textbf{c}_{j}$, which is optimized (along with the full transformer network) by taking the final loss of the probability of each correct answer $p_{i^{*}}$ over all answer choices, i.e., $\mathcal{L} = \sum_{d \in |D|} -\log p^{(d)}_{i^{*}}$. 


We specifically use \textbf{BERT-large} uncased with whole-word masking, as well as the \textbf{RoBERTa-large} model from \citet{liu2019roberta}, which is a more robustly trained version of the original BERT model. Our system uses the implementations provided in AllenNLP \cite{gardner2018allennlp} and Huggingface \cite{wolf2019huggingfaces}.




\paragraph{Baselines and Sanity Checks.} 
When creating synthetic datasets, it is important to ensure that systematic biases, or \emph{annotation artifacts} \cite{gururangan2018annotation}, are not introduced into the resulting probes and that  the target datasets are sufficiently challenging (or \emph{good}, in the sense of \citet{hewitt2019designing}). To test for this, we use several of the MCQA baseline models first introduced in \citet{mihaylov2018can}, which take inspiration from the LSTM-based models used in \citet{conneau2017supervised} for NLI and various \emph{partial-input} baselines based on these models.

Following \citet{mihaylov2018can}'s notation, for any sequence $s$ of tokens in $\{ q^{(j)}, a_{1}^{(j)},...,a_{N}^{(j)}\} \in D$, an encoding of $s$ is given as the following: 
\begin{align*}
  h_{s}^{(j)} = \textbf{BiLSTM}(\textsc{embed}(s)) \in \mathbb{R}^{|s| \times 2h},
\end{align*}
where $h$ is the dimension of the hidden state in each directional network, and \textsc{embed}$(\cdot)$ assigns a token-level embeddings to each token in $s$\footnote{As in \citet{mihaylov2018can}, we experiment with using both \textbf{GloVe} \cite{pennington2014glove} and \textbf{ELMo} \cite{peters2018deep} pre-trained embeddings for \textsc{embed}.}. A contextual representation for each $s$ is then built by applying an element-wise \texttt{max} operation over $h_{s}$ as follows:
\begin{align*}
    r^{(j)}_{s} = \texttt{max}(h_{s}^{(j)}) \in \mathbb{R}^{2h} 
\end{align*}

With these contextual representations, different baseline models can be constructed. For example, a \textbf{Choice-Only} model, a variant of the well-known \emph{hypothesis-only} baseline used in NLI \cite{poliak2018hypothesis}, scores each choice $c_{i}$ in the following way: $\alpha_{i}^{(j)} = \textbf{W}^{T} r^{(j)}_{c_{i}} \in \mathbb{R}$
for $\textbf{W}^{T} \in \mathbb{R}^{2h}$ independently of the question and assigns a probability to each answer $p_{i}^{(j)} \propto e^{\alpha_{i}^{(j)}}$. 

A slight variant of this model, the \textbf{Choice-to-choice} model, tries to single out a given answer choice relative to other choices by scoring all choice pairs $\alpha_{i,i'}^{(j)} = \textsc{Att}(r^{(j)}_{c_{i}},r^{(j)}_{c_{i'}}) \in \mathbb{R}$ using a learned attention mechanism \textsc{Att} and finding the choice with the minimal similarity to other options (for full details, see their original paper). In using these partial-input baselines, which we train directly on each target probe, we can check whether systematic biases related to answer choices were introduced into the  data creation process.


A \textbf{Question-to-choice} model, in contrast, uses the contextual representations for each question and individual choice and an attention model \textsc{Att} model to get a score $\alpha^{(j)}_{q,i} = \textsc{Att}(r^{(j)}_{q},r^{(j)}_{c_{i}}) \in \mathbb{R}$ as above. Here we also experiment with using \textbf{ESIM} \cite{chen2017lstm} to generate the contextual representations for $q,c_{i}$ (which includes token-wise attention), as well as a \textbf{VecSimilarity} model that measures the average (cosine) vector similarity between question and answer tokens: $\alpha^{(j)}_{q,i} = \textsc{Sim}(\textsc{embed}(q^{(j)}),\textsc{embed}(c^{(j)}_{i}))$. These sets of baselines, \added{which have been shown to be weak on other benchmark MCQA tasks}, are primarily used \added{not as competitive models but to} check for artifacts between questions and answers that are not captured in the partial-input baselines. \added{This helps} ensure that the overall MCQA \added{probing} tasks are sufficiently difficult.

\subsection{Inoculation and Pre-training}
\label{sec:inoculation}

Using the various models introduced above, we train these models on benchmark tasks in the science domain and look at model performance on our probes with and without additional training on samples of probe data, building on the idea of \emph{inoculation} from \citet{liu2019inoculation}. Model inoculation is the idea of continuing to train models on new challenge tasks (in our cases, separately for each probe) using only a small amount of examples. Unlike in ordinary fine-tuning, the goal is not to learn an entirely re-purposed model, but to improve on (or \emph{vaccinate} against) particular phenomena (e.g., our synthetic probes) that potentially deviate from a model's original training distribution. 

Following a variant proposed by \citet{richardson2020probing}, for each pre-trained (science) model and architecture $M_{a}$ we continue training the model on  $k$ new probe examples (with a maximum of $k=$ 3000) under a set of hyper-parameter configurations $\{ 1, ..., J\}$ and identify, for each $k$, the model $M_{*}^{a,k}$ with the best aggregate performance $S$ on the original (\emph{orig}) and \emph{new} task: 
\begin{align*}
    M^{a,k}_* & = \!\!\!\!\!\!\! \argmax_{M \in \{M^{a,k}_1, \ldots, M^{a,k}_J\}}\!\!\! \textsc{avg}\bigg(S_\text{new}(M), S_\text{orig}(M)\bigg)
\end{align*}
As in \citet{richardson2020probing}, we performed comprehensive hyper-parameter searches that target especially learning rates and \# training iterations.

Using this methodology, we can see how much exposure to new data it takes for a given model to master a new task, and whether there are phenomena that stress particular models (e.g., lead to catastrophic forgetting of the original task). Given the restrictions on the number of fine-tuning examples, our assumption is that when models are able to maintain good performance on their original task during inoculation, \emph{the quickness with which they are able to learn the inoculated task provides evidence of prior competence}, which is precisely what we aim to probe.   To measure past performance, we define a model's \textbf{inoculation cost} as the difference in the performance of this model on its original task before and after inoculation, which serves as a \emph{control} on the target QA model. 

We pre-train on an aggregated training set of all benchmark science exams in Table~\ref{tab:benchmarks}.\footnote{To save space, we do not report scores for each individual science dataset, but we did verify that our best models achieve results comparable to the state of the art for each dataset.}


\begin{table}[ht]
    \centering
    \scalebox{0.82}{
    \begin{tabular}{l|c|c}
        \hline 
         \multicolumn{1}{c}{\textbf{Science Datasets}} & \multicolumn{1}{c}{\textbf{\#Questions}} & \textbf{$N$}  \\ \hline
         OpenBookQA \citealt{mihaylov2018can} & 4,957  & 4 \\ 
         SciQ \citealt{welbl2017crowdsourcing} & 11,675 & 4 \\ 
         TextBookQA \citealt{kembhavi2017you} & 7,611 & 4/5 \\ 
         ARC Dataset++ \citealt{clark2018think} & 4,035 & 4/5 \\ \cdashline{1-3}
         MCQL \citealt{liang2018distractor} & 6,318 & 4 \\ \cdashline{1-3}
         \textbf{Science Collection} (total) & 34,596 & 5 \\ \hline 
    \end{tabular}}
    \caption{The MCQA training datasets used. \textbf{\#Question} denotes the number of training samples in our version of each dataset, $N$ the number of choices.}
    \label{tab:benchmarks}
\end{table}

In line with our goal of obtaining insights into the strongest QA models, we first pre-trained our \textbf{RoBERTa}-large model on the RACE dataset \cite{lai2017race}, a recipe used by several leading models on science benchmarks. and created an aggregate development set of $\sim$4k science questions for evaluating overall science performance and inoculation cost. To handle a varying number of answer choices in these sets, we made all sets 5-way by adding empty answers as needed. We also experimented with a slight variant of inoculation, called \textbf{add-some inoculation}, which involves balancing the inoculation training sets with naturalistic science questions. We reserve the MCQL dataset in Table~\ref{tab:benchmarks} for this purpose, and experiment with balancing each probe example with one science example (\emph{x1 matching}) and adding twice as many science questions (\emph{x2 matching}, up to 3k) for each new example. 

\begin{table*}[t]
    \centering
    \scalebox{0.72}{
    \begin{tabular}{ l c c c c c }
        \hline 
        & \multicolumn{4}{c}{\textbf{WordNetQA}} & \multicolumn{1}{|c}{\textbf{DictionaryQA}} \\ 
          & \textbf{Definitions} & \textbf{Synonymy} & \textbf{Hypernymy} & \textbf{Hyponymy} & \multicolumn{1}{|c}{\textbf{Word sense}} \\ 
         \textbf{Model} & \emph{(Dev/Test)} & \emph{(Dev/Test)} & \emph{(Dev/Test)} & \emph{(Dev/Test)} & \multicolumn{1}{|c}{\emph{(Dev/Test)}} \\ \hline 
         \multicolumn{6}{c}{Group 1: \textbf{Baselines} (direct training on 3k probes)} \\ 
\textbf{Random} & 19.9 / 20.0 & 19.8 / 19.8 & 19.9 / 20.0 & 20.2 / 21.0 & 20.0 / 19.0 \\
\textbf{Choice-Only}-GloVe & 26.6 / 26.1 & 36.9 / 36.1 & 42.5 / 46.0 & 34.3 / 34.4 & 35.0 / 32.1 \\
\textbf{Choice-Only}-BERT & 22.9 / 23.2 & \underline{\textbf{41.1}} / 39.4 & \underline{\textbf{63.8}} / 54.4 & 35.7 / 35.1 & 36.6 / 31.7 \\
\textbf{Choice-Only}-RoBERTa & \underline{\textbf{26.8}} / \underline{\textbf{28.6}} & 40.9 / \underline{\textbf{40.1}} & 62.3 / \underline{\textbf{57.3}} & \underline{\textbf{37.8}} / \underline{\textbf{37.5}} & \underline{\textbf{38.0}} / \underline{\textbf{31.7}} \\
\textbf{Choice-to-Choice}-GloVe & 26.4 / 28.1 & 40.1 / 35.0 & 47.0 / 35.5 & 35.4 / 36.1 & 37.3 / 33.3 \\
         \cdashline{1-6}
\textbf{Question-to-Choice}-VecSimilarity & 33.4 / 32.1 & 31.7 / 30.7 & 28.9 / 33.0 & 26.2 / 28.8 & 29.5 / 33.1 \\
\cdashline{1-6}
        \multicolumn{6}{c}{Group 2: \textbf{Task-Specific (non-transformer) Models}} \\ 
\textbf{Question-to-Choice}-GloVe & \underline{\textbf{53.6}} / \underline{\textbf{51.8}} & 57.3 / 55.3 & 50.4 / 47.0 & \underline{\textbf{61.6}} / \underline{\textbf{64.2}} & \underline{\textbf{53.2}} / \underline{\textbf{53.5}} \\
\textbf{Question-to-Choice}-ELMO & 42.3 / 41.6 & \underline{\textbf{58.6}} / \underline{\textbf{56.0}} & \underline{\textbf{56.0}} / \underline{\textbf{51.5}} & 54.8 / 56.3 & 51.6 / 52.1 \\
         \hline 
         \multicolumn{6}{c}{Group 3: \textbf{Science Models} (no fine-tuning or direct training on probes)} \\
\textbf{ESIM}-GloVe & 27.5 / 28.3 & 25.1 / 26.1 & 27.0 / 33.0 & 23.6 / 24.8 & 31.9 / 32.5 \\
\textbf{ESIM}-ELMO & 23.1 / 24.0 & 21.1 / 21.5 & 27.1 / 32.7 & 18.0 / 18.5 & 28.3 / 31.5 \\
         \cdashline{1-6}
\textbf{BERT} & 54.1 / 55.7 & 58.8 / 60.9 & 43.2 / 51.0 & 24.0 / 27.0 & 43.0 / 42.9 \\
\textbf{RoBERTa} & \underline{\textbf{74.1}} / \underline{\textbf{77.1}} & \underline{\textbf{61.1}} / \underline{\textbf{64.2}} & \textbf{\underline{53.2}} / \textbf{\underline{71.0}} & \underline{\textbf{48.5}} / \underline{\textbf{58.6}} & \underline{\textbf{53.0}} / \underline{\textbf{55.1}} \\
         \hline 
         \multicolumn{6}{c}{Group 4: \textbf{Science Models} (best aggregate model $M_{*}$ fine-tuned on probes; \textbf{inoculation cost} is shown in parenthesis)} \\ 
\textbf{ESIM}-GloVe &  46.2 / 42.4 (-6.27)   &  50.4 / 47.3 (-6.84)   &  56.6 / 52.9 (-5.69)   &  59.1 / 61.1 (-5.10)   &  50.0 / 55.3 (-7.09)  \\
         \cdashline{1-6}
\textbf{BERT} &  84.0 / 84.1 (-1.15)   &  79.6 / 79.7 (-0.44)   &  73.8 / 82.7 (-0.49)   &  79.8 / 88.0 (-0.92)   &  75.6 / 79.1 (-2.84)  \\
\textbf{RoBERTa} &  \underline{\textbf{89.0}} / \underline{\textbf{89.3}} (-1.33)   &  \underline{\textbf{81.2}} / \underline{\textbf{81.3}} (-1.31)   &  \underline{\textbf{77.7}} / \underline{\textbf{87.7}} (-0.74)   &  \underline{\textbf{81.2}} / \underline{\textbf{89.4}} (-1.64)   &  \underline{\textbf{80.0}} / \underline{\textbf{85.9}} (-2.23)  \\
         \hline 
\T \textbf{Human Performance} (estimates) & -- / 91.2\% \ \ \ \ & -- / 87.4\% \ \ \ \ & -- / 96\%$^\dagger$ \ \ \ \ \ & -- / 95.5\%$^\dagger$ \ \ \ \ \ & -- / 95.6\%$^\dagger$ \ \ \ 
    \end{tabular}}
    
    \caption{\textbf{Instance-level} accuracy (\%) of all baselines (group 1), task-specific non-transformer QA models (group 2), pre-trained MCQA models (zero-shot, group 3), and MCQA models after fine-tuning on our probes (group 4). Human scores marked with $^\dagger$ represent scores on gold-standard annotated test sets.}
    \label{tab:instance}
\end{table*}





\begin{figure*}[t]
    \centering
    \begin{tabular}{p{5.8cm} p{4.2cm} p{4.5cm}}
    
    
    \multicolumn{1}{c}{\scriptsize \textbf{ESIM+GloVe-Science (no-training)}} & \multicolumn{1}{c}{\scriptsize \textbf{ESIM+GloVe-Science (100 ex.)}} & \multicolumn{1}{c}{\scriptsize \textbf{ESIM+GloVe-Science (3000 ex.)}} \\[0cm]
      \includegraphics[scale=.33]{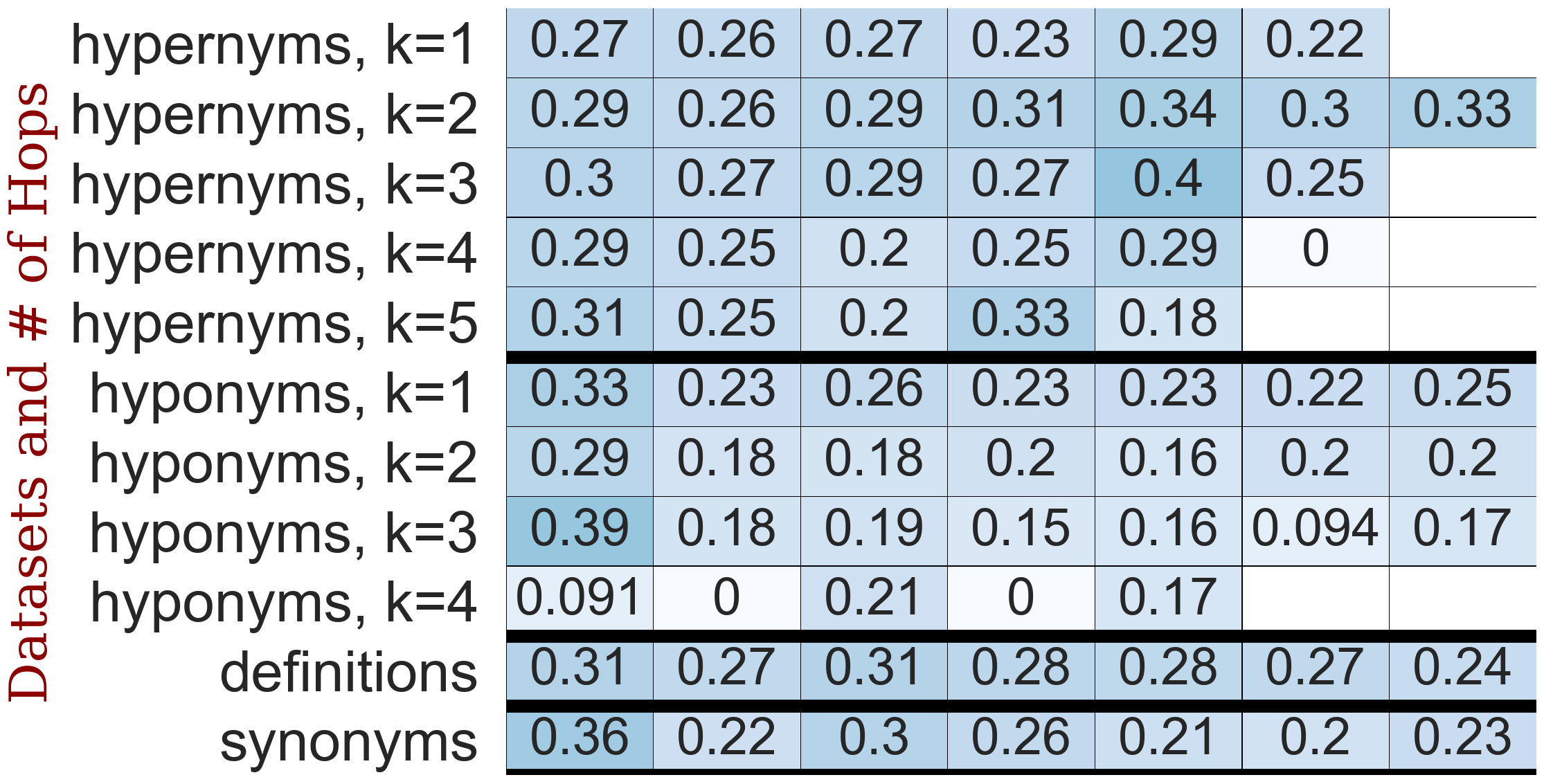}   & \includegraphics[scale=.33]{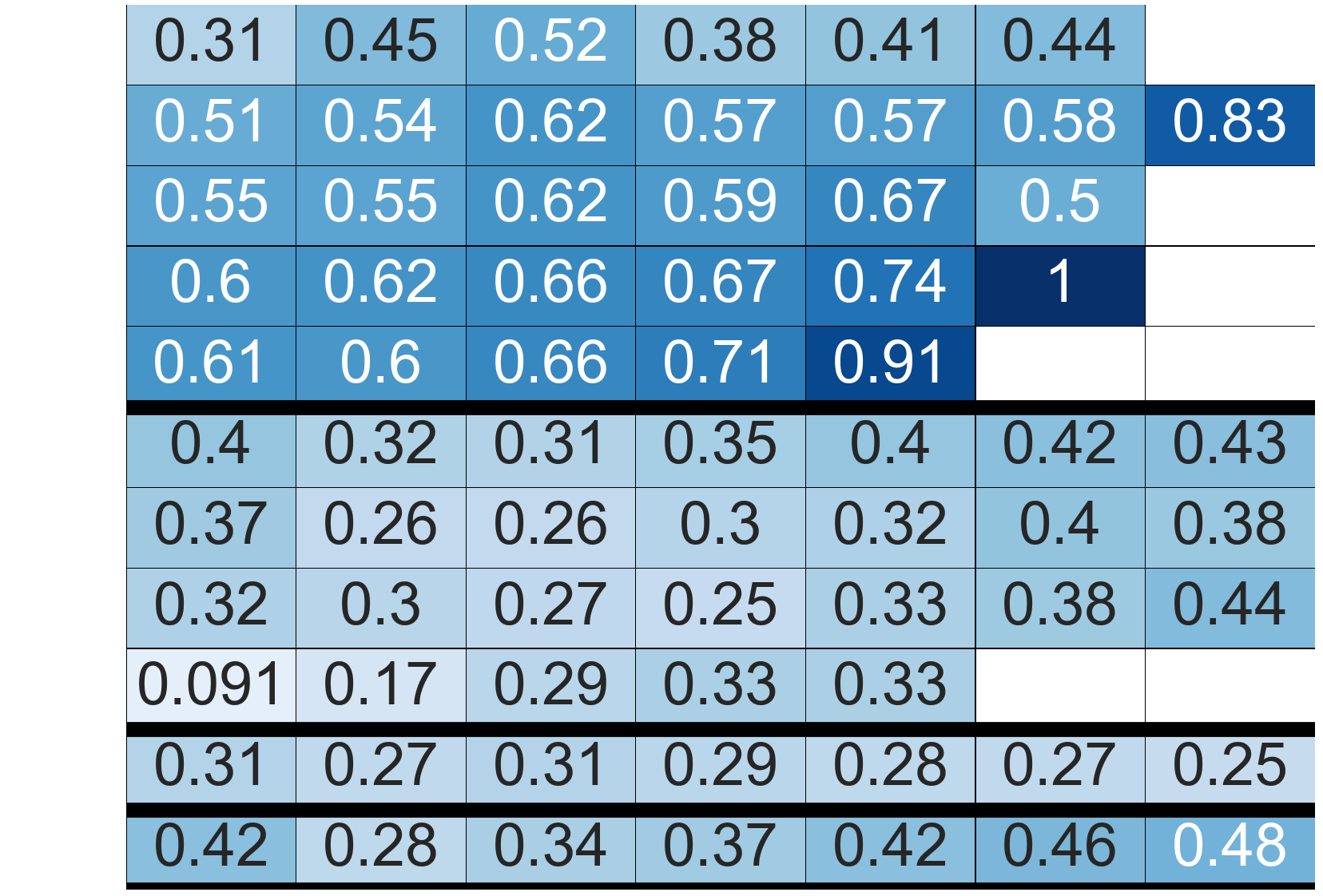} & \includegraphics[scale=.33]{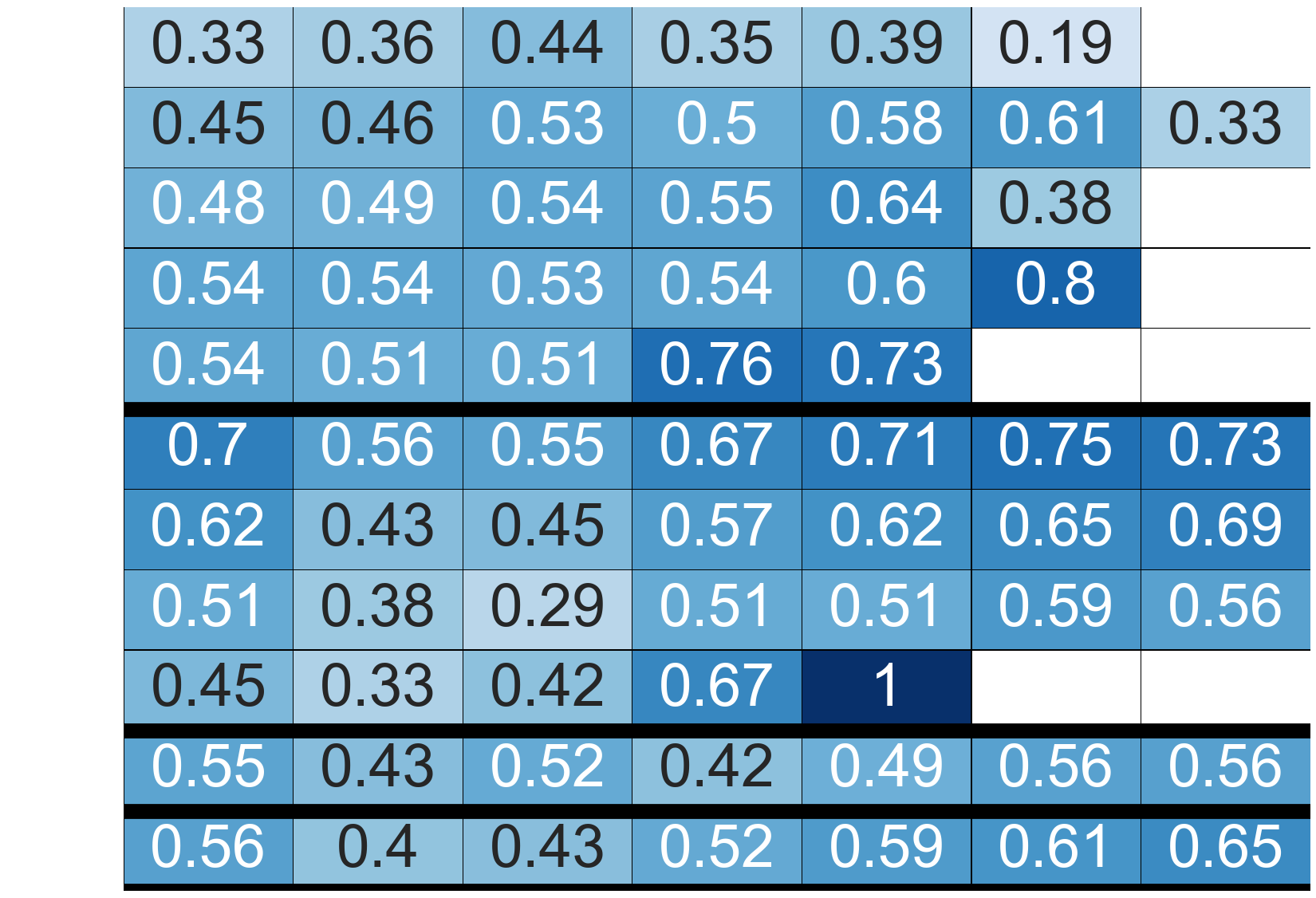} \\[-.2cm]

    
    
    \multicolumn{1}{c}{\scriptsize \textbf{RoBERTa-Science (no-training)}} & \multicolumn{1}{c}{\scriptsize \textbf{RoBERTa-Science (100 ex.)}} & \multicolumn{1}{c}{\scriptsize \textbf{RoBERTa-Science (3000 ex.)}} \\[0cm]
    \includegraphics[scale=.33]{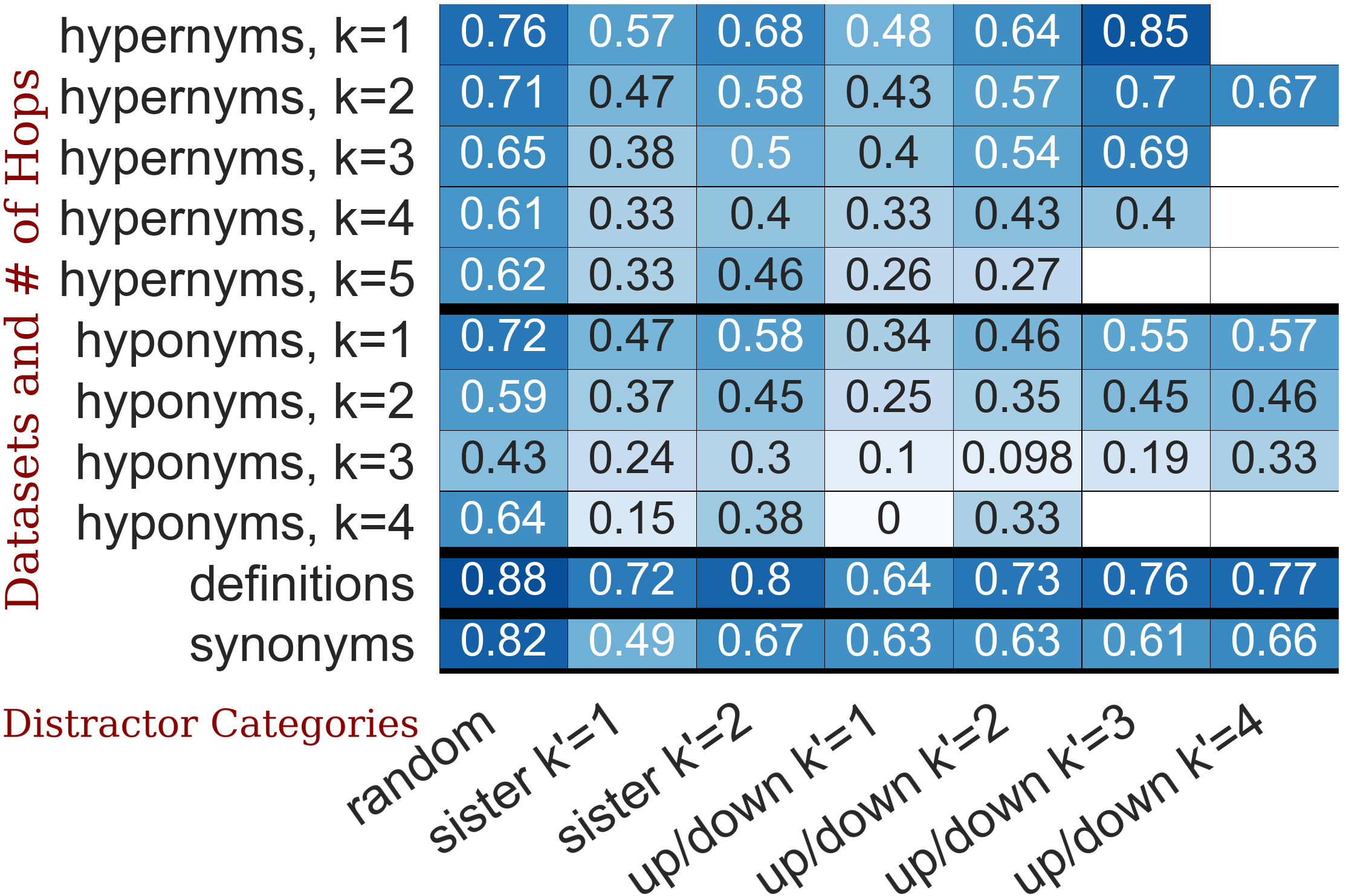} & \includegraphics[scale=.33]{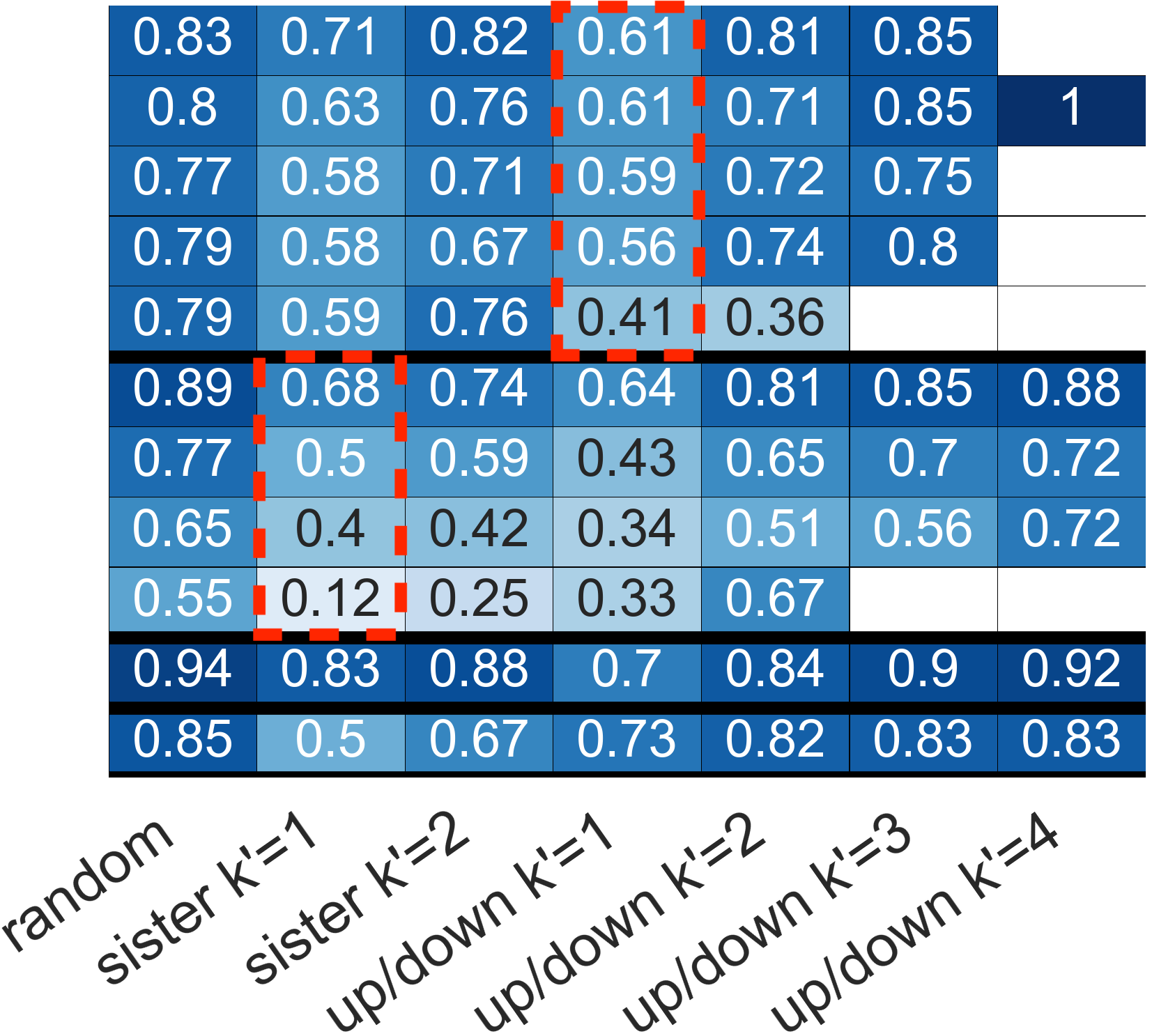} & \includegraphics[scale=.33]{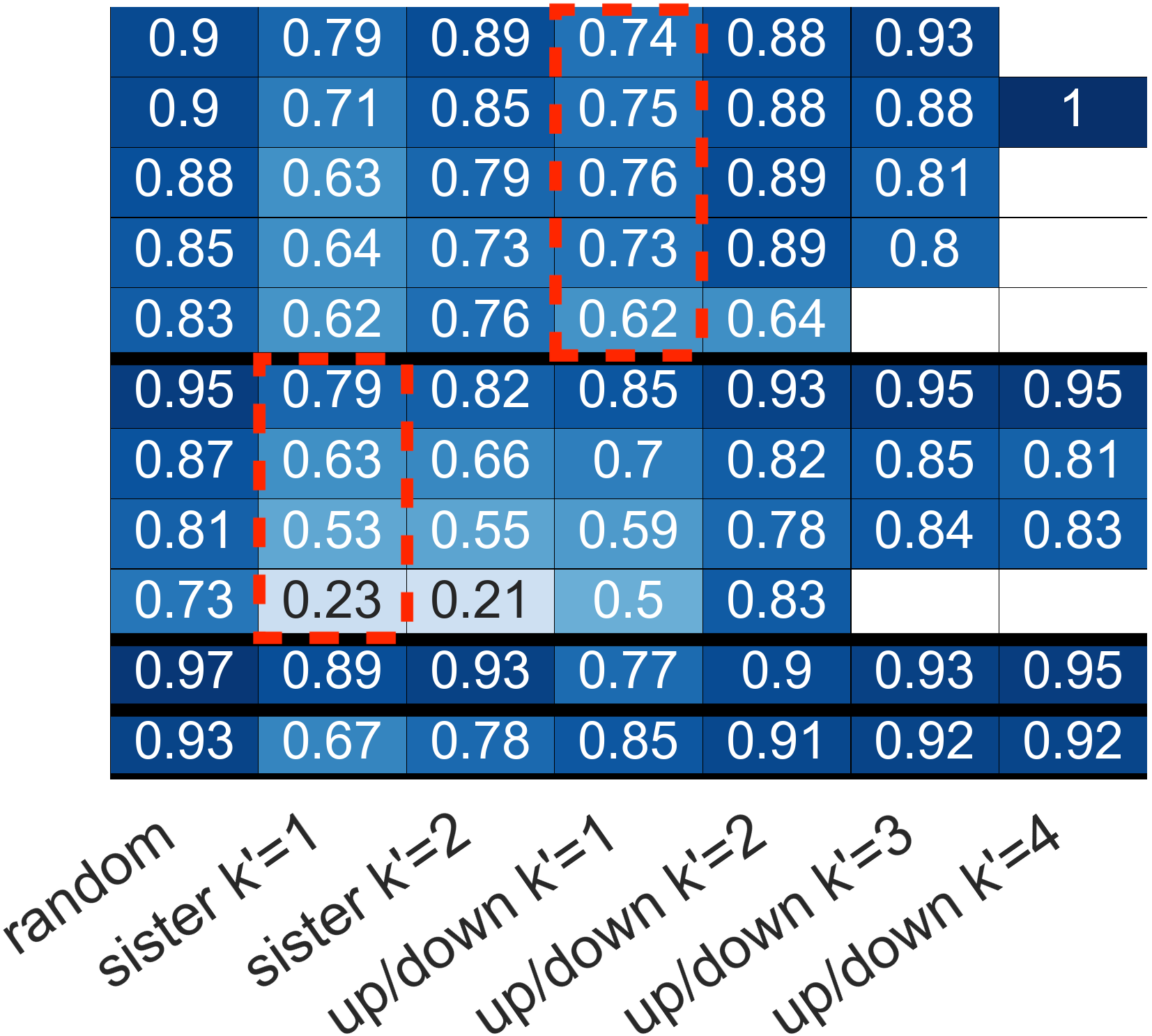} \\[-.3cm]
    
    \end{tabular}
    \caption{Combined model accuracies on the different WordNetQA datasets (divided by 4 bold lines) broken down (where possible) into number of hops $k$ (rows) and types of distractor sets and hops $k'$ (rows) across the different stages of inoculation (\textbf{\# ex.}). The 4 dashed lines show some trends related to multi-hop inference.}
    \label{fig:distractor_study}
\end{figure*}

\subsection{Evaluating Model Competence} 
\label{sec:eval}

We use \textbf{instance-level accuracy}, the standard overall accuracy of correct answer prediction (as in Table~\ref{tab:instance}). In addition, we also propose to measure a model's \textbf{cluster-level} (or \emph{strict cluster}) \textbf{accuracy}, which requires correctly answering all questions in a \emph{semantic cluster} (cf.~Section~\ref{sec:clusters}).

Our cluster-based analysis is motivated by the idea that if a model truly knows the meaning of a given concept then it should be able to answer arbitrary questions about this concept without sensitivity to varied distractors. While our strict cluster metric is simplistic, it takes inspiration from work on visual QA \cite{shah2019cycle}, and allows us to evaluate a model's \emph{consistency} and \emph{robustness} across our different probes, and to get insight into whether errors are concentrated on a small set of concepts or widespread across different clusters.

\added{The ability of a model to answer several questions about a single concept can be thought of as a type of \emph{certificate} (i.e., further justification and demonstration) of general understanding of that concept in the sense of \citet{ranta2017explainable}.}

\section{Results and Findings}
\label{sec:results_findings}

\added{We begin with an assessment to ensure that our probes are sufficiently difficult to provide meaningful insights into strong models (Section~\ref{subsec:are-probes-challenging}), then assess the strength of pre-trained QA models (Section~\ref{subsec:are-pretrained-models-strong}) and whether they can be effectively inoculated (Section~\ref{subsec:model-inoculation}), and finally present a cluster-based consistency analysis (Section~\ref{subsec:cluster-consistency}).}

\subsection{Are our probes sufficiently challenging?}
\label{subsec:are-probes-challenging}
Partial-input baseline models, \textbf{Choice-Only} and \textbf{Choice-to-Choice}, generally performed poorly on our probes (cf.~Table~\ref{tab:instance}, group 1), indicating limited biases in distractor generation. Initial versions of DictionaryQA had unforeseen biases partly related to distractors sampled from entries without example sentences (cf.~Section~\ref{sec:dqa}), which resulted in high (56\%) Choice-Only-GloVe scores before such distractors were filtered out.

One exception is our hypernymy probe where, despite several attempts at filtering data and de-duplicating splits (w.r.t.\ correct answer and distractor types), the Choice-to-Choice-BERT/RoBERTa models achieve over 60\% accuracy. The nature of the biases here remains unclear, highlighting the importance of having rigorous baselines as unintended biases in expert knowledge can carry over to resulting datasets. We also note the large gap between the BERT/RoBERTa versus GloVe choice-only models, emphasizing the need for using the best available models even in partial-input baselines.

A more conventional set of \emph{Task-Specific} QA models (i.e., the LSTM-based \textbf{Question-to-Choice} models trained directly on the probes) is not particularly strong on any of the datasets (cf.~Table~\ref{tab:instance}, group 2), suggesting our probes are indeed sufficiently challenging and largely immune from overt artifacts. The poor performance of the \emph{VecSimilarity} (which uses pre-trained Word2Vec embeddings without additional training) provides additional evidence of the insufficiency of elementary lexical matching strategies.

\subsection{How strong are pre-trained QA models?}
\label{subsec:are-pretrained-models-strong}
Non-transformer science models, such as \textbf{ESIM} with GloVe or ELMo, struggle with all probes (cf.~Table~\ref{tab:instance}, group 3), often scoring near random chance. In sharp contrast, the transformer models have mixed results, the most striking being \textbf{RoBERTa} QA models on the definitions, synonymy and hypernymy test  probes (achieving 77\%, \added{64}\%, and 71\% resp.), which substantially outperform even task-specific LSTM models trained directly on the probes. Throughout all of these results, however, model performance is significantly behind human performance.

At first glance, these zero-shot results suggest RoBERTa's high competence on these phenomena. A closer scrutiny enabled by our controlled probes, however, provides a more subtle picture. Each heat map in Figure~\ref{fig:distractor_study} breaks down the performance of an ESIM or RoBERTa QA model based on the difficulty of the probe dataset (rows) and the nature of the distractors (columns).

Across all datasets and number of hops in the question (i.e., all rows), zero-shot model performance for RoBERTa (bottom-left heat map) is consistently highest among examples with random distractors (the first column) and lowest when distractors are closest in WordNet space (e.g., sister and ISA, or \emph{up/down}, distractors at distance $k'=1$). For example, RoBERTa's zero-shot score drops from 88\% to 64\% when going from random distractors to \emph{up/down} distractors at $k'=1$.

Further, model performance also clearly degrades for hypernymy and hyponymy as $k$, the number of hops in the question, increases (see red dashed boxes). For example, the accuracy on questions involving hyponym reasoning with sister distractors of $k'=1$ (column 2) degrades from 47\% to only 15\% as $k$ increases from 1 to 4. This general tendency persists despite additional fine-tuning, providing evidence of the limited ability of these models to perform multi-hop inference.

\subsection{Can models be effectively inoculated?}
\label{subsec:model-inoculation}
How well do probe generation templates align with the science training distribution (which we know little about) can significantly impact zero-shot performance~\cite{petroni2019language}. Zero-shot results above thus provide a \emph{lower bound} on model competence on the probed phenomena. We next consider a probe-specific fine-tuning or \emph{inoculation} step, allowing models to learn target templates and couple this with knowledge acquired during pre-training and science training. 

Accuracy after inoculation on 3K probe instances is shown (with inoculation cost in parenthesis) in group 4 of Table~\ref{tab:instance}, for the model with the highest aggregate score on the original task and new probe. Transformer-based models again outperform non-transformer ones, and \emph{better models correlate with lower inoculation costs}. E.g., on synonymy, ESIM's inoculation cost is 7\%, but only $\sim$1\% for BERT and RoBERTa.
This emphasizes the high capacity of transformer QA models to absorb new phenomena at minimal cost, as observed earlier for NLI~\cite{richardson2020probing}. 

\begin{figure}[t]
    \centering
    \begin{tabular}{p{3.2cm}  p{3.5cm}}
    \advance\leftskip-.24cm
    \includegraphics[scale=.25]{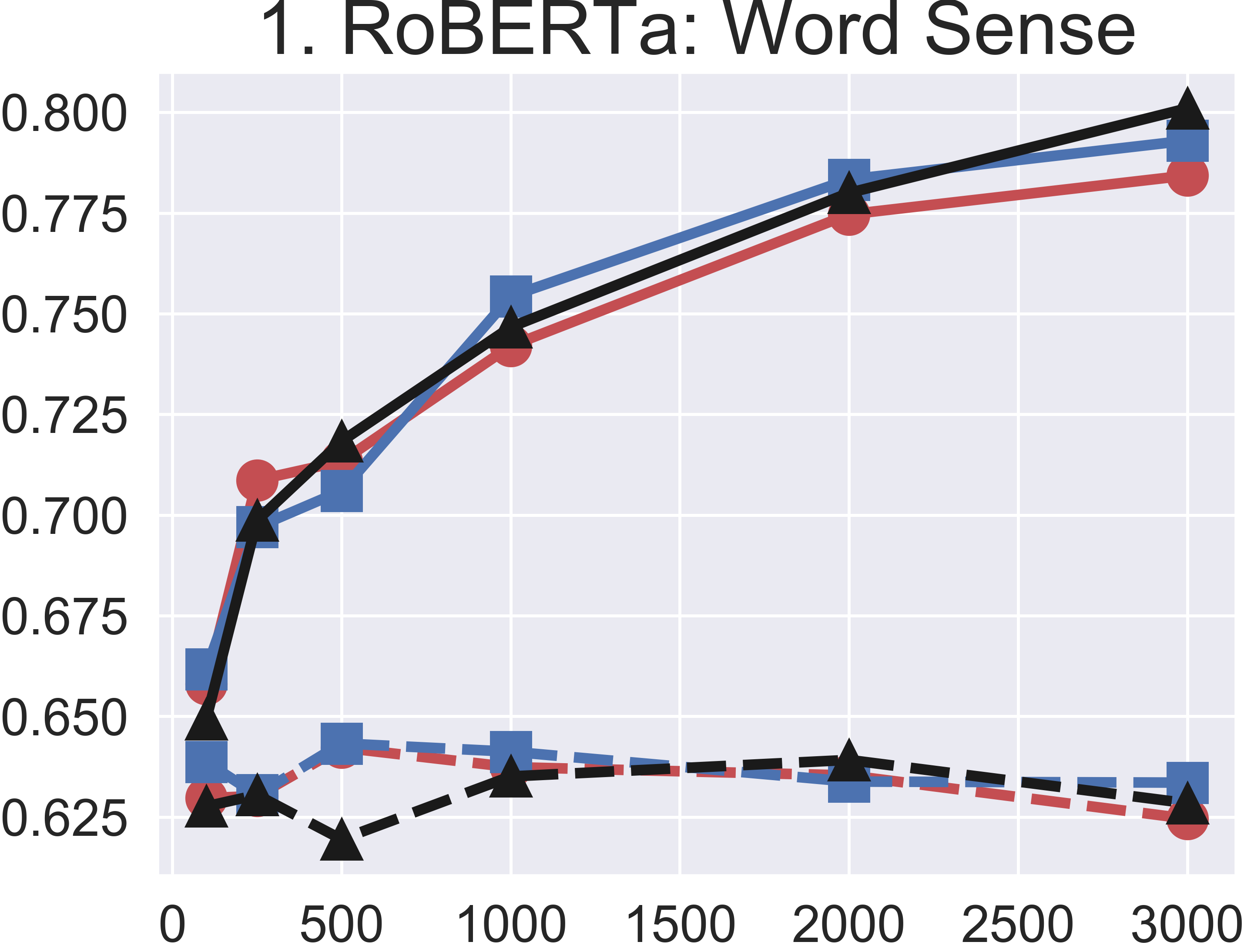} & 
    \includegraphics[scale=.25]{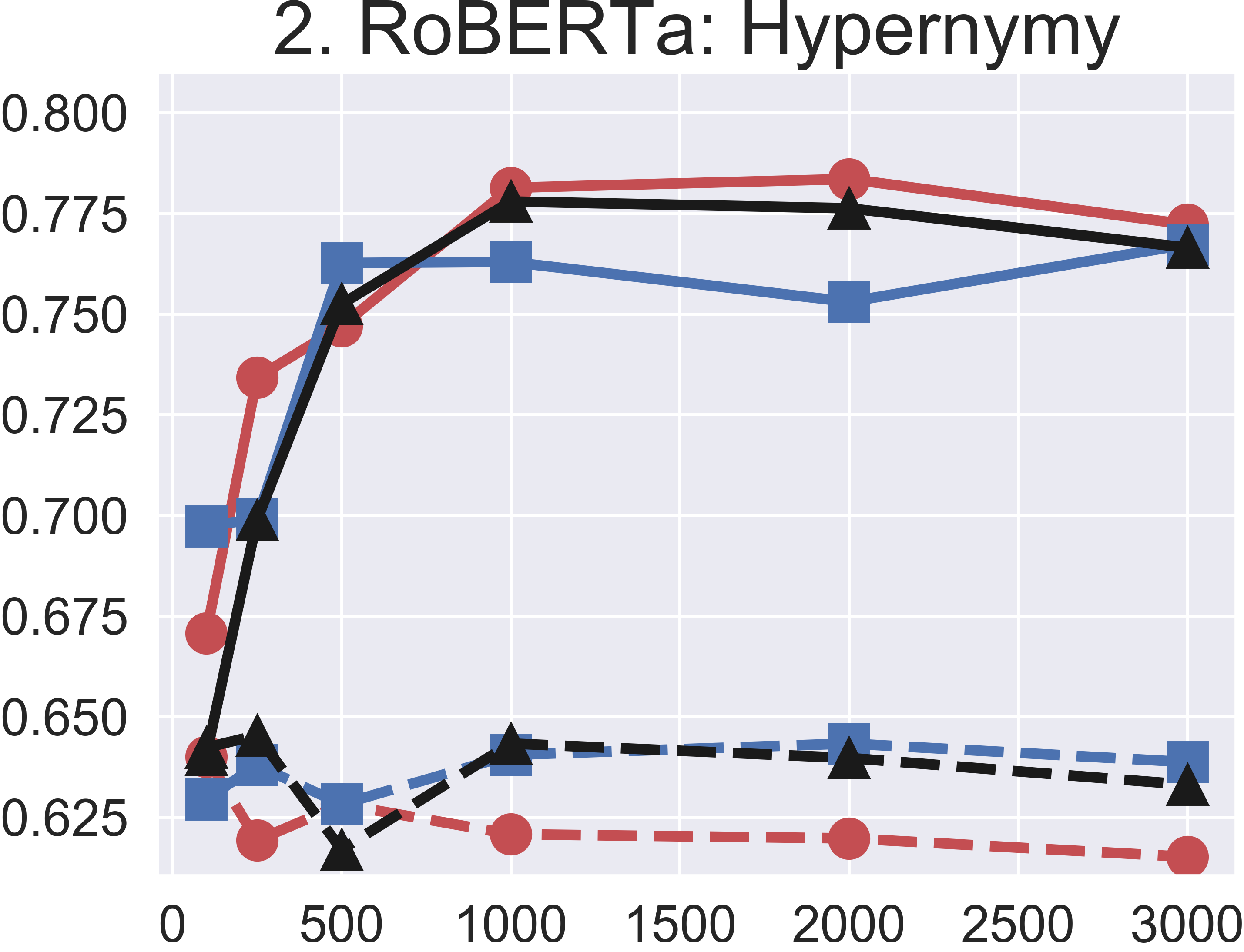}
    \\
    \advance\leftskip-.24cm
    \includegraphics[scale=.25]{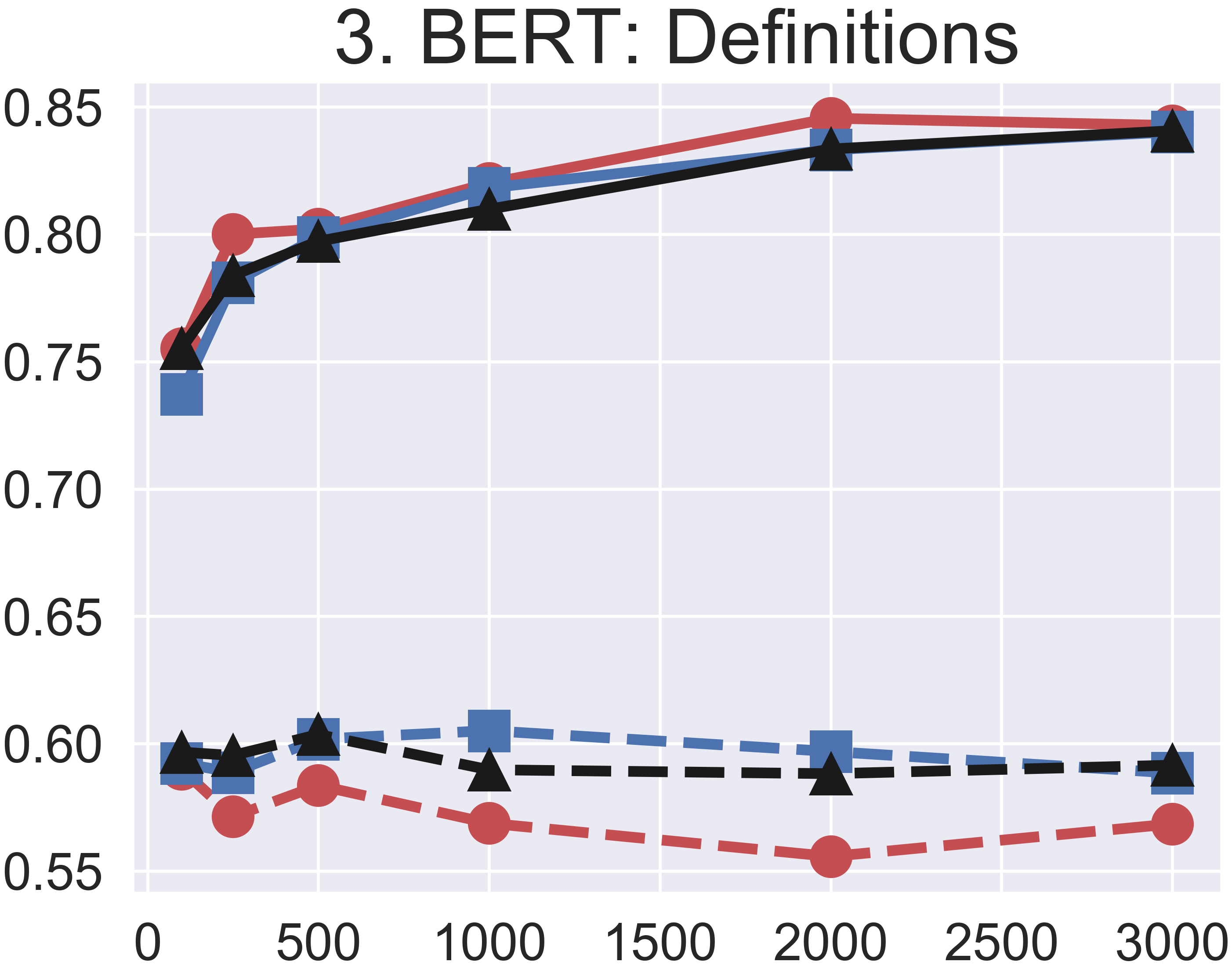}  & 
    \includegraphics[scale=.25]{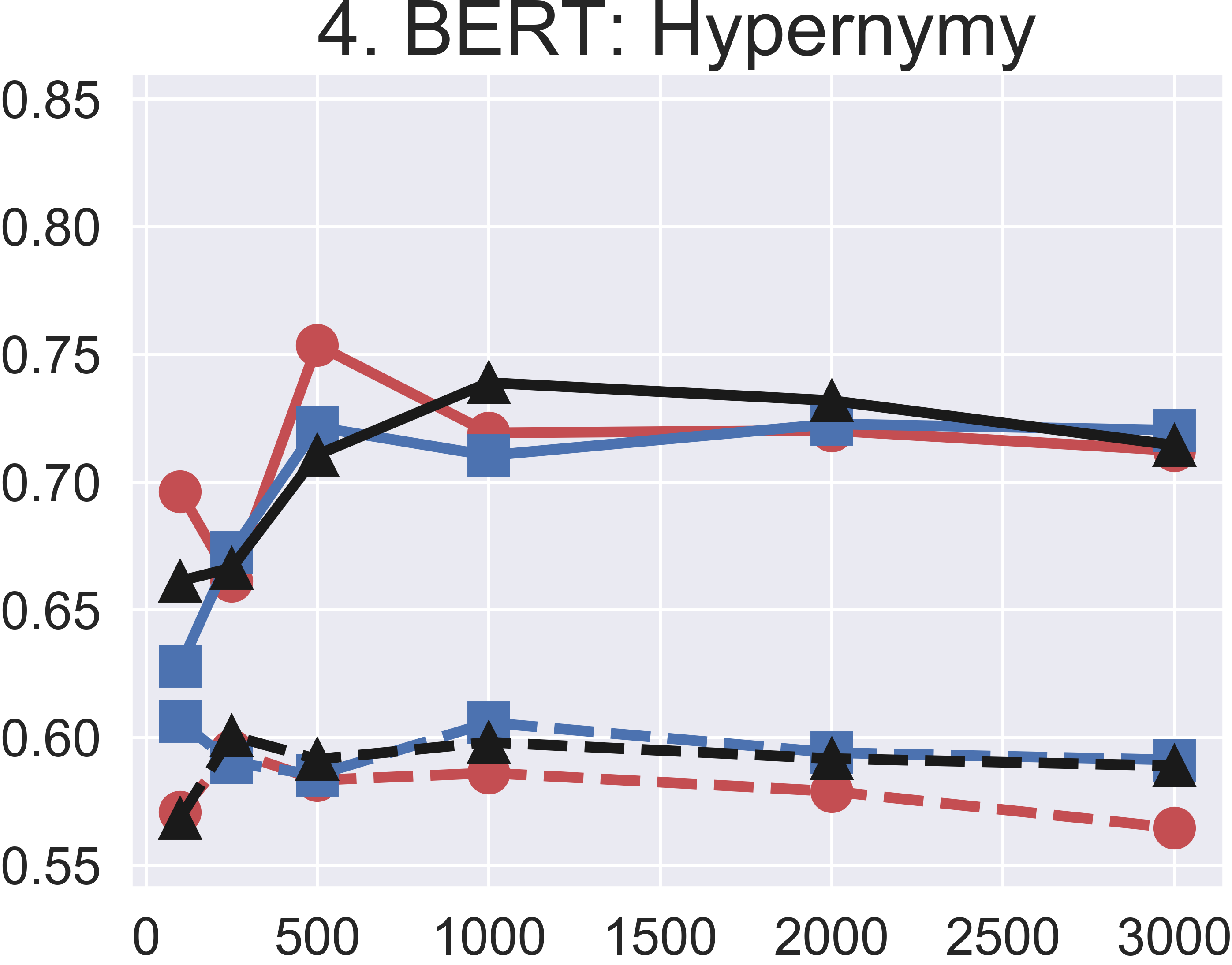}
    \\
    \advance\leftskip-.24cm
    \includegraphics[scale=.25]{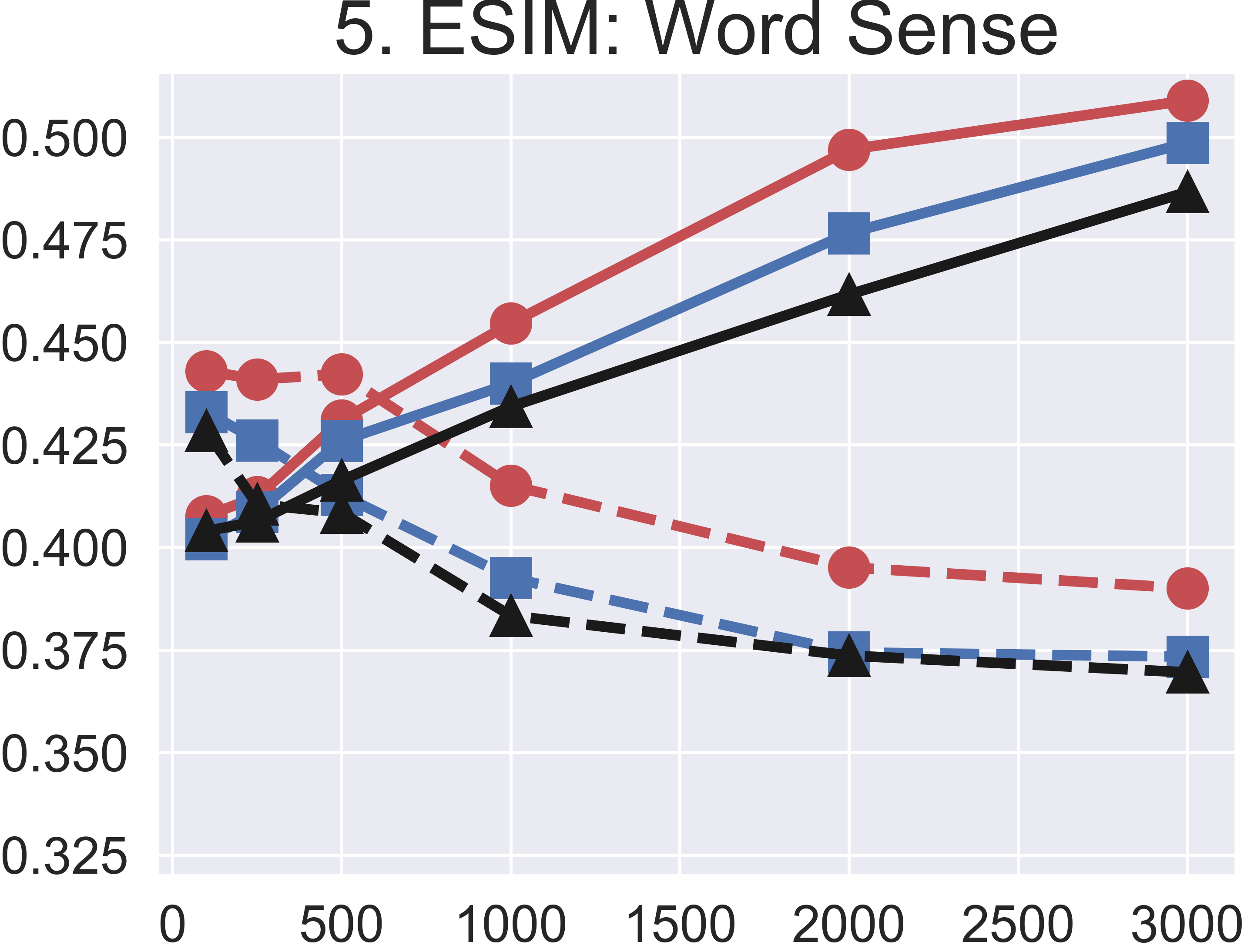} & 
    \includegraphics[scale=.25]{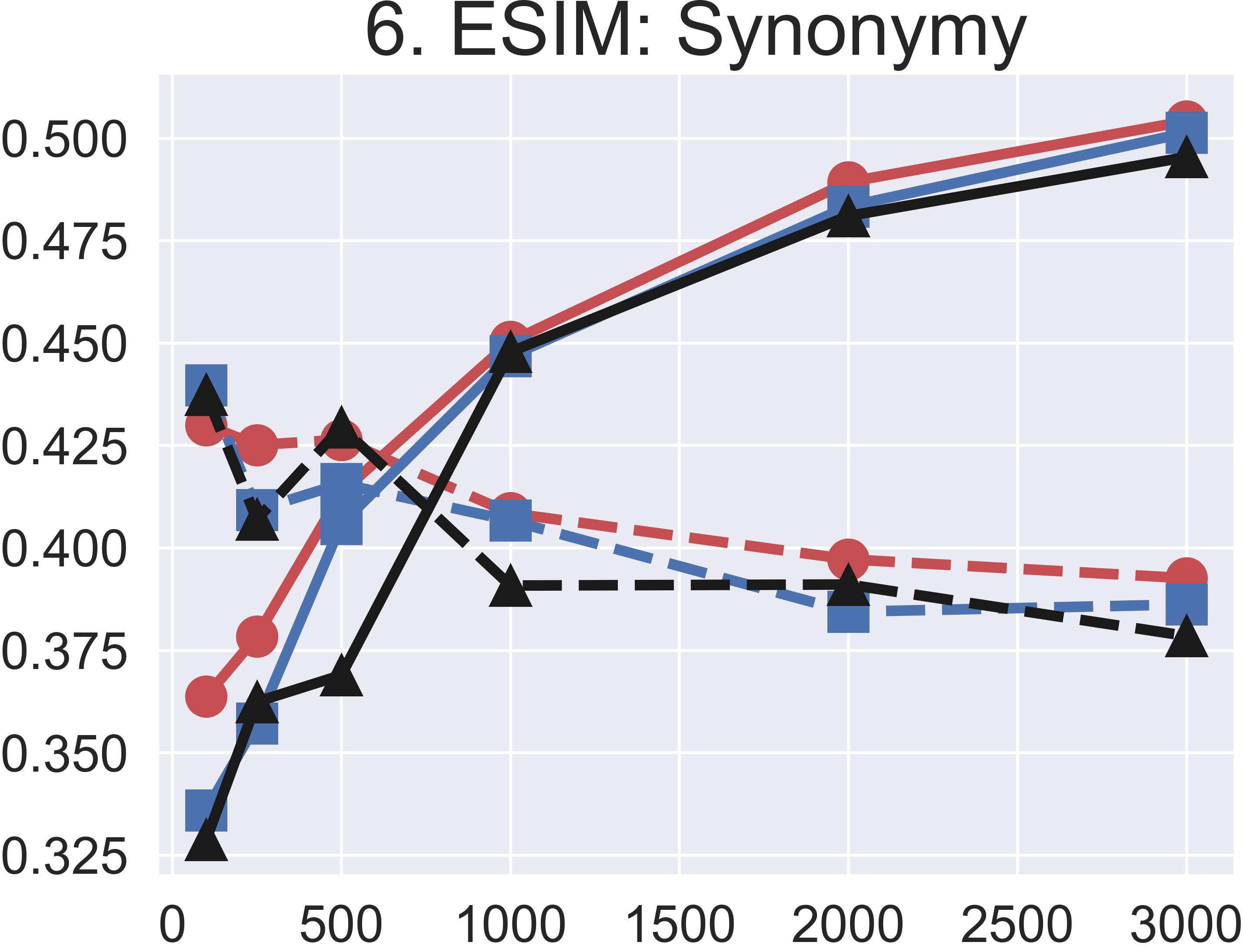} \\[-.2cm]
    \end{tabular}
    \caption{
        Inoculation plots with accuracy on challenge tasks (red/circle solid lines) and original tasks (red/circle dashed lines) using the best aggregate model $M_{*}^{a,k}$ at each $k$ challenge examples (x axis). The effect of using \textbf{add-some inoculation} is shown in the blue/square (\emph{x1 match}) and black/triangle (\emph{x2 match}) lines.
    }
    \label{fig:inoculation_plots}
\end{figure}

Figure~\ref{fig:inoculation_plots} shows the corresponding learning curves. Transformer QA models learn most tasks quickly while maintaining constant scores on their original tasks (flat dashed lines, plots 1-4), providing evidence of high competence. For BERT and RoBERTa, \textbf{add-some inoculation} (a) improves scores on the probing tasks (\emph{solid} black and blue lines, plot 1) and (b) minimizes loss on the original task (\emph{dashed} blue and black lines, plots 2-4).

ESIM behaves quite the opposite (plots 5-6), generally unable to learn individual probes without degrading on its original task. More science data during inoculation confuses it on both tasks.

As the middle-bottom plot of Figure~\ref{fig:distractor_study} shows, RoBERTa's performance improves significantly (e.g., from 59\% to 77\% on 2-hop hyponymy with random distractors) even after inoculation with a mere 100 examples, providing strong evidence of prior competence. After 3k examples, it performs well on virtually all probes. However, results still notably degrade with the number of hops and distractor complexity, as discussed earlier, and we still find its performance to be between 2\%-10\% behind human performance.

\subsection{Are models consistent across clusters?}
\label{subsec:cluster-consistency}
Table~\ref{tab:cluster_results_simple} shows mixed results for \textbf{cluster-level} accuracy across the different WordNetQA probes. Our best model is rather robust on the definitions probe. RoBERTa QA's cluster accuracy is $75\%$, meaning it can answer \emph{all} questions correctly for 75\% of the target concepts, and that errors are concentrated on a small minority (25\%) of concepts. On synonymy and hypernymy, both BERT and RoBERTa are less strong but appear robust on a majority of concepts. In contrast, our best model on hyponymy has an accuracy of only 36\%,
indicating that the RoBERTa QA models knows only partially about a vast majority of concepts, leaving substantial room for further improvement.

\added{We emphasize that these results only provide a crude look into model consistency and robustness. Recalling dataset details in Table~\ref{tab:wordnetqa}, probes differ in terms of the average size of clusters. For example, hyponymy, in virtue of having many more questions per cluster, might simply be a much more difficult dataset for our cluster-based evaluation. In addition, such a strict evaluation does not take into account potentially erroneous questions within clusters, which is an important issue that we leave for future work.}


\begin{table}[t]
    \centering
    \scalebox{0.65}{
    \begin{tabular}{l c c c c}
        \hline 
         & \textbf{Definitions} & \textbf{Synonymy} & \textbf{Hypernymy} & \textbf{Hyponymy}  \\
         \textbf{Model} & \multicolumn{4}{c}{\emph{Strict Cluster Accuracy} ($\Delta$)} \\ \hline
         \textbf{Choice-Only} & 14.7 (-12.0) & 18.5 (-22.3) & 34.6 (-27.6) & 4.1 (-33.7) \\ \cdashline{1-5}
         \textbf{ESIM} & 30.2 (-15.9) & 23.3 (-26.9) & 29.2 (-27.3) & 15.2 (-43.8) \\
\textbf{BERT} & 68.5 (-15.5) & 58.1 (-21.5) & 49.0 (-24.8) & 34.0 (-45.4) \\
\textbf{RoBERTa} & 75.0 (-13.9) & 61.7 (-19.4) & 54.0 (-23.2) & 36.7 (-44.4) \\
    \end{tabular}}
    \caption{\textbf{Cluster-level} accuracies (\%) on the WordNetQA dev.~sets for inoculated models and best \textbf{Choice-only} model. $\Delta$ show the absolute difference in percentage points with instance-level accuracies.}
    \label{tab:cluster_results_simple}
\end{table}


\section{Discussion}
\label{sec:discussion}

We presented a new methodology for automatically building challenge datasets from knowledge graphs and taxonomies. We introduced several new silver-standard datasets for systematically probing state-of-the-art open-domain QA models. While our focus was on probing definitions and ISA reasoning, the methodology is amendable to any target knowledge resource or QA domain. We see synthetic datasets and our general methodology as an inexpensive supplement to recent large-scale investment in \emph{naturalistic} QA dataset construction \cite{zellers2018swag,sakaguchi2019winogrande} to help better understand today's models.

We found transformer-based QA models to have a remarkable ability to reason with complex forms of relational knowledge, both \emph{with} and \emph{without} exposure to our new tasks. In the latter case (zero-shot), a newer RoBERTa QA model trained only on benchmark data outperforms several \emph{task-specific} LSTM-based models trained directly on our probes. When \emph{inoculated} using small samples (e.g., 100 examples) of probing data, RoBERTa masters many aspects of our probes with virtually no performance loss on its original QA task---which we use as a control on the probing quality. 

Since these models seem to already contain considerable amounts of relational knowledge, our simple inoculation strategy, which nudges models to bring out this knowledge explicitly while retaining performance on their original task (hence allowing a fairer probe of its knowledge by giving the model the opportunity to learn the probe format), could serve as a simpler alternative to designing new model architectures explicitly encoding such knowledge \cite{peters2019knowledge}. 

Regarding our focus on preserving a model performance on its original task, one might expect that re-training on relevant knowledge should \emph{improve} performance. Following other work in this area \cite{richardson2020probing,yanaka2020neural}, we found that maintaining performance after additional fine-tuning on specialized datasets is already a tall order given that models are susceptible to over-specialization; indeed, similar issues have been noticed in recent work on large-scale transfer learning \cite{raffel2019exploring}. We believe that using inoculation for the sole purpose of improving model performance, which is beyond the scope of this paper, would likely require a more sophisticated inoculation protocol. Devising more complex loss functions extending our inoculation strategy to help balance old and new information could help in this endeavor.

The main appeal of automatically generated probes is the ability to systematically manipulate probe complexity, which in turn enables more controlled experimentation as well as new forms of evaluation. It allowed us to study in detail the effect of different types of distractors and the complexity of required reasoning. This study showed that even the best QA models, despite additional fine-tuning, struggle with harder categories of distractors and with multi-hop inferences. For some probes, our cluster-based analysis revealed that errors are widespread across concept clusters, suggesting that models are not always consistent and robust. These results, taken together with our findings about the vulnerability of synthetic datasets to systematic biases and comparison with human scores, suggest that there is much room for improvement and that the positive results should be taken with a grain of salt. Developing better ways to evaluate semantic clusters and model robustness would be a step in this direction.

We emphasize that using synthetic versus naturalistic QA data comes with important trade-offs. While we are able to generate large amounts of systematically controlled data at virtually no cost or need for manual annotation, it is much harder to validate the quality of such data at such a scale and such varying levels of complexity. \added{Conversely, with benchmark QA datasets, it is much harder to perform the type of careful manipulations and cluster-based analyses we report here.} While we assume that the expert knowledge we employ, in virtue of being hand-curated by human experts, is generally correct by design, we know that such resources are fallible and error-prone. We propose measuring human performance via small samples of probing data, and leave more scalable methods of removing potential noise and adding human annotation to future work. 

\added{One of the overarching goals of our approach to model probing is to uncover whether black box models are able to reason in a consistent and correct manner. Our assumption, similar to \citet{clark2020transformers}, is that the ability of a model to mimic the input-output behavior of data generated using expert knowledge gives some evidence of correctness in virtue of such data being \emph{correct by construction} (see discussion by \citet{ranta2017explainable}). We emphasize, however, that there are limits to how much we can learn through this type of behavioral testing, given that models are susceptible to exploiting systematic biases in synthetic data and the general difficulty of disentangling a model's knowledge acquired during pre-training versus fine-tuning \cite{talmor2019olmpics}. We therefore see efforts to combine behavioral testing with various other \emph{analysis methods} \cite{belinkov2019analysis} that aim to uncover correlations and causal patterns between internal model representations and discrete structures \cite{chrupala2019correlating,vig2020causal,geiger2020modular} as a promising direction for future work. This, in combination with extending our probing strategy to other forms of expert knowledge, could prove to be an effective way to engage others working on linguistics and other areas of AI in state-of-the-art NLP research.}




\section*{Acknowledgments}

\added{We thank the Action Editor and the three anonymous reviewers for their thoughtful comments and feedback. Thanks also to our colleagues at AI2, in particular Peter Clark, Daniel Khashabi, Tushar Khot, Oyvind Tafjord, and Alon Talmor for feedback on earlier drafts of this work and assistance with various aspects of modeling. Special thanks to Daniel Khashabi for helping with some of the earlier human evaluation experiments.}

\bibliographystyle{acl_natbib}
\begin{small}
\bibliography{natlog}
\end{small}


\end{document}